\documentclass[10pt,twocolumn,letterpaper]{article}

\usepackage{cvpr}              % To produce the CAMERA-READY version
\usepackage{graphicx}
\usepackage{amsmath}
\usepackage{amssymb}
\usepackage{booktabs}

\usepackage{times}
\usepackage{epsfig}
\usepackage{graphicx}
\usepackage{amsmath}
\usepackage{amssymb}
\usepackage{booktabs}
\usepackage{multirow}
\usepackage{pifont}
\usepackage{enumitem}

\usepackage[pagebackref,breaklinks,colorlinks]{hyperref}
\usepackage[capitalize]{cleveref}
\crefname{section}{Sec.}{Secs.}
\Crefname{section}{Section}{Sections}
\Crefname{table}{Table}{Tables}
\crefname{table}{Tab.}{Tabs.}

\def\cvprPaperID{7675} % *** Enter the CVPR Paper ID here
\def\confName{CVPR}
\def\confYear{2022}

\newcommand{\nocaps}{\texttt{nocaps}}

\begin{document}

%%%%%%%%% TITLE
% \title{}
\title{
UFO: A UniFied TransfOrmer for Vision-Language Representation Learning
}
% Scaling Vision-Language Transformer \\
% VLVL: A transformer for Vision, Language and Vision-Language \\
% SIT: A SIngle Transformer for Vision-Language Representation Learning \\

\author{Jianfeng Wang, Xiaowei Hu, Zhe Gan, Zhengyuan Yang, \\
Xiyang Dai, Zicheng Liu, Yumao Lu, Lijuan Wang\\
Microsoft \\
{\tt\small \{jianfw,Xiaowei.Hu,zhe.gan,zhengyang,xiyang.dai,zliu,yumaolu,yumaolu\}@microsoft.com}
}

\maketitle

\begin{abstract}
In this paper, we propose a single UniFied transfOrmer (UFO), which is capable of processing either unimodal inputs (e.g., image or language) or multimodal inputs (e.g., the concatenation of the image and the question), for vision-language (VL) representation learning.
Existing approaches typically design an individual network for each modality and/or a specific fusion network for multimodal tasks.
To simplify the network architecture, we use a single transformer network and enforce multi-task learning during VL pre-training, which includes the image-text contrastive loss, image-text matching loss, and masked language modeling loss based on the bidirectional and the seq2seq attention mask.
The same transformer network is used as the image encoder,
the text encoder, or the fusion network in different pre-training tasks.
Empirically, we observe less conflict among different tasks and achieve new state of the arts on visual question answering, COCO image captioning (cross-entropy optimization) and \nocaps~(in SPICE). 
On other downstream tasks, e.g., image-text retrieval, we also achieve competitive performance.

\end{abstract}

\section{Introduction}
\begin{figure}
    \centering
    \includegraphics[width=0.99\linewidth]{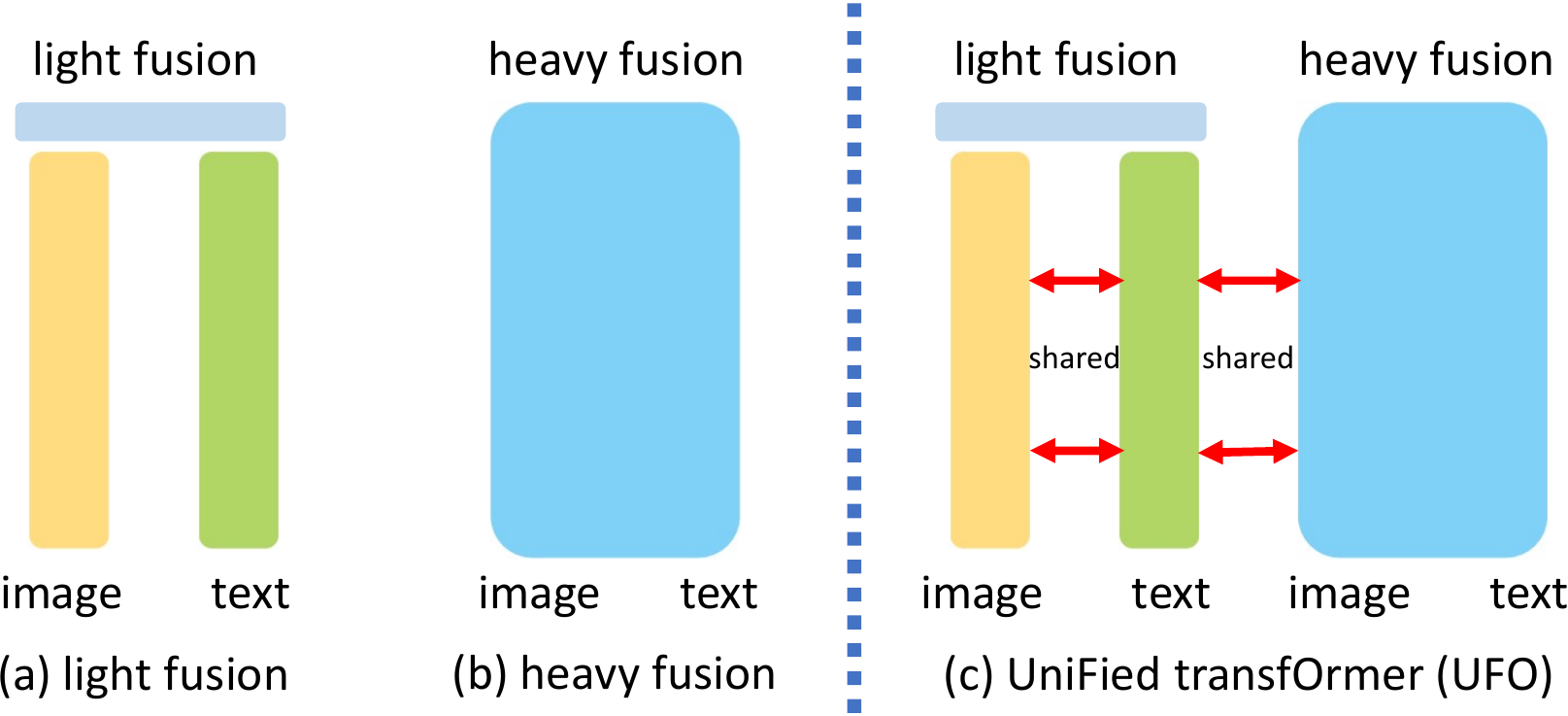}
    % \vspace{1mm}
    \caption{Different network designs for VL representation learning. 
    (a) Light fusion: few parameters are dedicated to multimodal fusion. 
    (b) Heavy fusion: a transformer network is applied to fuse the multimodal inputs. 
    Before fusion, the image can be encoded as region features, patch features, or CNN grid features, and the text can be encoded through an embedding layer or a transformer network.
    (c) Our UniFied transfOrmer (UFO): a single network is reused as image encoder, text encoder, and fusion network for different tasks.
    }
    \label{fig:light_heavy}
    % \vspace{-2mm}
\end{figure}

Recent years have seen tremendous progress in vision-language (VL) representation learning, where the model is designed to understand the vision/language signals and the relation between the modalities.
Applications include image captioning~\cite{LinMBHPRDZ14,abs-1812-08658}, visual question answering (VQA)~\cite{GoyalKSBP16}, image-text retrieval, \etc.
Typical approaches first extract the features from each modality and 
then feed them to a fusion network to jointly learn the representation. 
Regarding how the fusion network is designed, we roughly group the existing approaches into two categories: light fusion and heavy fusion, as shown in Fig.~\ref{fig:light_heavy}(a) and (b).

Light fusion (Fig.~\ref{fig:light_heavy}(a)) adopts a separate encoder for both the image and the text, such as in CLIP~\cite{RadfordKHRGASAM21} and ALIGN~\cite{JiaYXCPPLSLD21}.
The image encoder can be ResNet~\cite{HeZRS16} or 
vision transformer~\cite{DosovitskiyB0WZ21}, while the text encoder is typically a transformer network~\cite{VaswaniSPUJGKP17}.
The fusion is the contrastive loss based on the cosine similarity 
such that the representation from the two modalities can be aligned into the same semantic space.
A favorable application is the image-text retrieval, where each image or text description is represented as a fixed vector for fast similarity search.

While few parameters are allocated in the light fusion, heavy fusion approaches,
shown in Fig.~\ref{fig:light_heavy}(b), applies a transformer network on top of the unimodal features to jointly learn the representation.
The image can be encoded as object features \cite{00010BT0GZ18,Li0LZHZWH0WCG20,ChenLYK0G0020,ZhouPZHCG20,LuBPL19,SuZCLLWD20,TanB19,abs-1908-03557,LiDFGJ20,Li0LZHZWH0WCG20,abs-2009-13682,abs-2101-00529,abs-2012-06946,abs-2104-02096}
through a Faster RCNN~\cite{RenHGS15},
as grid features \cite{jiang2020defense,abs-2004-00849} from a convolutional neural network,
or as patch features \cite{KimSK21} from a linear projection on raw pixels.
The text can be encoded as token representations by a transformer network as in~\cite{TanB19,abs-2107-07651}
or by a simple embedding layer as in~\cite{Li0LZHZWH0WCG20,abs-2101-00529,abs-2012-06946,abs-2009-13682,KimSK21}.
With a heavier fusion network, the final representation can better capture the contextual connection between the modalities. 
A typical application is the VQA, where the network predicts the answer based on both the image and the question.

Existing approaches design different network architectures for different tasks.
As the transformer network can be used for all these components, in this paper we attempt to design a single UniFied transfOrmer (UFO) for both light-fusion and heavy-fusion scenarios, as shown in Fig.~\ref{fig:light_heavy}(c).
For the light-fusion tasks, the transformer is used as both the image encoder and 
the text encoder. 
For the heavy-fusion tasks, the same transformer is reused as a fusion module to process the two signals together.

Before feeding into the transformer, the raw image pixels are grouped into patches and projected by a linear mapping. 
The text description is projected into the same dimension by an embedding layer.
In this way, we allocate as few as possible learnable parameters in the 
modality-specific processing and devote the majority in the shared transformer network. 
The network is adjusted automatically on how much representation power should be allocated for each individual modality and how much for the joint representation. 

To empower the network with the capability for unimodal inputs, 
we apply the image-text contrastive (ITC) loss on the outputs during VL pre-training (VLP).
For multimodal fusion capability, we apply the image-text matching (ITM) loss and the masked language modeling (MLM) loss based on the bidirectional and \texttt{seq2seq} attention.
% Intuitively, it is challenging to share one network for different roles, but we observe less conflict from different tasks
% empirically. 
To optimize the network with multiple tasks, we randomly choose one of the tasks in each iteration for efficiency and leverage a momentum teacher motivated by~\cite{abs-1911-05722,abs-2107-07651}
to guide the learning. 
With extensive ablation study, we observe less conflict among these tasks. In certain cases, different tasks even help each other. For example, the MLM task
improves the retrieval task based on ITC significantly. 
Meanwhile, we also achieve new state of the arts\footnote{As of 11/2021 among peer-reviewed publications.} in VQA, COCO image captioning (cross-entropy optimization), and \nocaps (in SPICE),
and competitive performance on other downstream tasks, \eg, image-text retrieval.
% In the meanwhile, the single pretrained transformer can achieve competitive accuracy compared with the individual design for each specific task
% after finetuning. 

\section{Related Work}
% As in Fig.~\ref{fig:light_heavy}, we categorize the existing work in light fusion and heavy fusion regarding how many learnable parameters are used for modality fusion.

\subsection{Light Fusion}
CLIP~\cite{RadfordKHRGASAM21} aligns both the image and the text description through a contrastive loss. 
The image is encoded by ResNet~\cite{HeZRS16} or vision transformer~\cite{DosovitskiyB0WZ21} and is represented as a single vector after a pooling operator. 
The text description is encoded by a transformer network and is also represented by a single vector.
During pre-training, the contrastive loss aligns the representations of the image-text pair to be similar, while 
the representations from mismatched image-text pairs to be dissimilar.
ALIGN~\cite{JiaYXCPPLSLD21} further scales up the contrastive loss on the large-scale noisy image-text data and mainly adopts 
EfficientNet~\cite{TanL19} as the image encoder.
Beyond the grid-level representation, LightningDOT~\cite{sun2021lightningdot} explores the region-level features through a Faster RCNN~\cite{RenHGS15} network to obtain the image representation.
A favorable application is the image-text retrieval. 
For zero-shot image classification, the image classes can be interpreted as a text description and the problem can be converted as a text retrieval task. 

VATT~\cite{abs-2104-11178} extends the contrastive learning from the image domain to the video domain and aligns the video frames, audios, and texts. A shared transformer network is also studied in~\cite{abs-2104-11178} as a modal-agnostic network. The difference with our work is that VATT~\cite{abs-2104-11178} belongs to the light-fusion, while we focus on unifying the light-fusion  and the heavy-fusion.

\subsection{Heavy Fusion}
Heavy-fusion networks allocate much more parameters in the modality fusion. Before fusion, each modality is encoded into the same dimensional space. 
For the image, one widely-adopted approach is to use an off-the-shelf 
Faster RCNN model~\cite{RenHGS15,00010BT0GZ18}
to extract multiple region features, \eg, as  in~\cite{00010BT0GZ18,Li0LZHZWH0WCG20,ChenLYK0G0020,ZhouPZHCG20,LuBPL19,SuZCLLWD20,TanB19,abs-1908-03557,LiDFGJ20,Li0LZHZWH0WCG20,abs-2009-13682,abs-2012-06946,abs-2104-02096}. 
VinVL~\cite{abs-2101-00529} explores an even stronger region detector to push the performance, while MiniVLM~\cite{abs-2012-06946} designs an efficient region detector for real-time VL applications. 
Instead of the region (sparse) features, 
another recent direction is to use the grid (dense) features, 
which can be extracted from ResNet as in~\cite{jiang2020defense,abs-2004-00849,abs-2104-12763},
or from a vision transformer as in~\cite{abs-2107-07651,xue2021probing}.
An advantage of the grid feature is that the image encoder can be trained or fine-tuned together
with other network components without the bounding box annotations. 
For the text, a simple approach is to apply one embedding layer after tokenization, such as the models in~\cite{Li0LZHZWH0WCG20,abs-2101-00529,abs-2012-06946,abs-2009-13682,KimSK21,ChenLYK0G0020}, or to leverage a specific transformer network as in~\cite{abs-2107-07651,TanB19}.

With the extracted features, the fusion network is applied to learn the contextual representation. 
The structure can be based on the cross-attention module between different modalities as in~\cite{TanB19,LuBPL19},
or on the self-attention module as in~\cite{ChenLYK0G0020,Li0LZHZWH0WCG20,abs-2101-00529,abs-2012-06946} 
where the features of multiple modalities are simply concatenated.
In~\cite{cho2021unifying,abs-2104-12763}, % simvlm: \cite{wang2021simvlm}
a transformer encoder and decoder are applied on the visual features and text prefix.
In~\cite{cho2021unifying}, multiple tasks are unified as a text generation process. This unification is still under the category of the heavy fusion. That is, the learned network cannot be used to process the unimodal inputs, \eg, image only for light-fusion applications. 
We attempt to unify in the network level, while~\cite{cho2021unifying} in the task level.

\section{UniFied TransfOrmer}
\begin{figure*}[t!]
    \centering
    \includegraphics[width=0.95\linewidth]{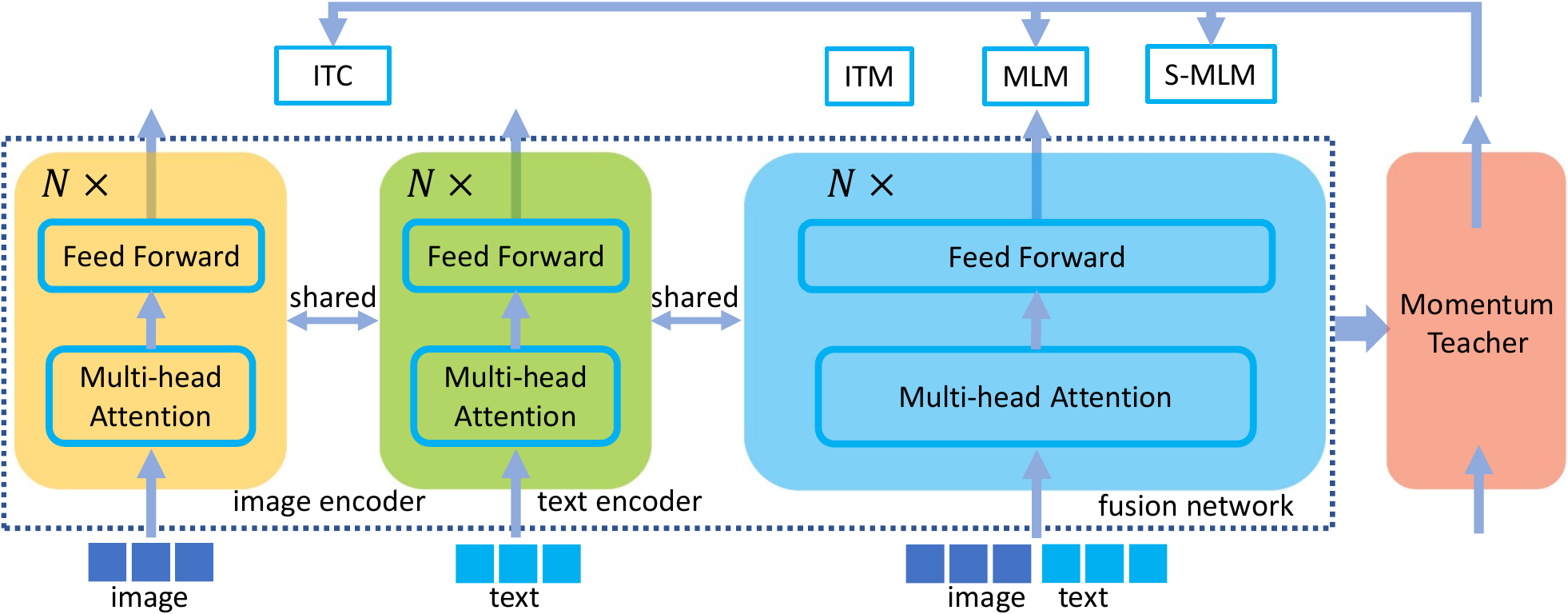}
    % \vspace{1mm}
    \caption{Vision-language pre-training of our UniFied transfOrmer (UFO).
    A single transformer is learnt to behave as an image encoder, a text encoder and a fusion network.
    The pre-training losses include the image-text contrastive (ITC) loss,
    image-text matching (ITM) loss,
    masked language modeling loss based on the bidirectional (MLM) and \texttt{seq2seq} attention mask (S-MLM).
    ITC empowers the network to understand the unimodal inputs (image or text),
    while the rest three focus on the joint inputs. 
    In each iteration, one of the losses is randomly selected and is guided by a momentum teacher if the loss is ITC/MLM/S-MLM.
    }
    \label{fig:framework}
    % \vspace{-2mm}
\end{figure*}
The key idea is to leverage only one transformer network, which is reused as an image encoder,
a text encoder, or a fusion encoder.
We follow the widely-used pretraining-then-finetuning scheme to train the network.
During the pre-training, we are given a large corpus of image-text pairs and multiple losses are enforced to empower the network for different roles. For unimodal signals, we apply the image-text contrastive loss~\cite{RadfordKHRGASAM21}. 
For multimodal fusion task, we apply 
the image-text matching loss (ITM)
and the masked language modeling loss based on both the bidirectional (MLM) and unidirectional (S-MLM) attention.
Fig.~\ref{fig:framework} illustrates the pre-training framework. 

\subsection{Network Structure}
We adopt the transformer network as the backbone. 
The main reason is that the transformer network has been demonstrated to perform well 
on the image tasks~\cite{DosovitskiyB0WZ21},
the language tasks~\cite{VaswaniSPUJGKP17}, and 
VL tasks~\cite{KimSK21,Li0LZHZWH0WCG20,abs-2101-00529,ChenLYK0G0020,abs-2009-13682,abs-2012-06946}.
Other choices are convolutional neural network (CNN) or
all-MLP~\cite{abs-2105-01601} structures, 
but it is unclear how to effectively apply such networks to all the three roles.
% Recent work~\cite{abs-2105-01601} investigates the pure MLP layers to
% replace the attention module and shows strong results. However, the network's parameter replies on the number of input tokens and thus is less appropriate to handle variant input lengths. Meanwhile, it is unclear how this all-MLP approach works on the multimodal tasks. 

The input to the transformer network is a sequence of tokens,
each of which is represented as a $d$-dimensional vector.
To tokenize the image, we split the raw image into disjoint patches,
each of which is linearly projected into the $d$-dimensional space with a learnable linear layer as in~\cite{DosovitskiyB0WZ21,KimSK21}. 
A learnable $2$-D positional embedding is added to each patch representation and an image \texttt{[CLS]} token is pre-appended.
The text description is first tokenized and then embedded to the $d$-dimensional space through an embedding matrix. 
A starting token of \texttt{[CLS]} and an ending token of \texttt{EOS} are added to wrap the text sequence.
A learnable $1$-D positional embedding is added to each text token.
Here, the image \texttt{[CLS]} and the text \texttt{[CLS]} are two different tokens. 
Before feeding the input to the transformer network, a modality-specific embedding is 
added to the corresponding input.
For VL tasks, the two modality inputs are concatenated before sending to the transformer. 
Although the inputs may have different token lengths,
the transformer network is naturally capable of handling variant input lengths.  

\subsection{Pre-training Tasks}
% The pretraining is performed on a large corpus of image-text pair data, \eg Conceptual Concept~\cite{SoricutDSG18}.

\paragraph{Image-Text Contrastive Loss.}
% Motivated by~\cite{RadfordKHRGASAM21,JiaYXCPPLSLD21}, 
% we apply the image-text contrastive (ITC) loss
% such that the transformer is able to process unimodal inputs.
The image-text contrastive (ITC) loss is to train the network to process either the image or the text and aligns the matched pairs 
into similar representations.
For the image, the network is used as an image encoder, and
the output corresponding to the \texttt{[CLS]} token is chosen
as the representation.
For the text, the network is reused as a text encoder, and 
the one corresponding to the \texttt{[EOS]} token is as the representation. 
The text \texttt{[CLS]}
is used for the image-text matching loss.
Let $\mathbf{I}_i$ and $\mathbf{T}_i$ be the $i$-th image and text representation ($l_2$-normalized), respectively.
Given $N$ pairs within a training batch, the loss is
\begin{align}
l_{\text{ITC}} & = \frac{1}{2} (l_1 + l_2), \\
l_1  & = 
    -\frac{1}{\sum_{i, j}\delta_{i, j} } \sum_{i, j} \delta_{i, j}\log \frac
    {
        \exp \{\mathbf{I}_i^T\mathbf{T}_j/t\}
    }
    {
        \sum_k \exp\{ \mathbf{I}_i^T\mathbf{T}_k / t \}
    }, \\
l_2 & =
    -\frac{1}{\sum_{i, j}\delta_{i, j} } \sum_{i, j} \delta_{i, j} \log \frac{
        \exp \{\mathbf{I}_i^T\mathbf{T}_j/t\}
    }{
        \sum_k \exp\{ \mathbf{I}_k^T\mathbf{T}_j / t \}
    },
\end{align}
where $t$ is initialized as 1 and is learnable as in~\cite{JiaYXCPPLSLD21}.
The indicator of $\delta_{i, t}$ is 1 if the $i$-th image is paired
with the $j$-th text, and $0$ otherwise. 
This is to handle the data where an image (or a text) is associated with multiple texts (or images), \eg, COCO~\cite{LinMBHPRDZ14}.
% Similar approach is also used in ALBEF~\cite{abs-2107-07651} for momentum-based contrastive learning but only for downstream tasks.

% \vspace{-4mm}
\paragraph{Image-Text Matching Loss.}%\label{sec:itm}
The network input is the concatenation of the matched
or mismatched image-text pairs. 
The mismatched pairs are constructed by randomly selecting a
text description in the dataset for a given image\footnote{In the implementation,
we randomly select the mismatched pairs from the current training batch to reduce the disk I/O.}.
The network is required to predict whether it is matched or not,
which is a binary classification task.
Here, we use the representation of the text \texttt{[CLS]} as the joint representation,
followed by a MLP layer for prediction. 
The cross-entropy loss is applied to penalize the incorrect prediction.

Motivated by~\cite{ChenLYK0G0020,KimSK21}, we apply the image-patch alignment loss. 
The image and text tokens are first connected through solving an
optimal transport problem by the inexact proximal point method~\cite{XieWWZ19}.
The resulting distance is maximized for mismatched image-text pairs and minimized for the matched pairs.
This loss is weighted by $0.1$ as in~\cite{ChenLYK0G0020,KimSK21}.
Overall, we use ITM to denote the sum of the two losses.

% \vspace{-2mm}
\paragraph{Masked Language Modeling Loss.}
The network input is the concatenation of the image tokens and the partially masked text tokens, while the transformer network is trained to predict the masked tokens.
As a common practice~\cite{DevlinCLT19}, 
$15\%$ of the text tokens are selected for prediction.
Each selected token is replaced with a \texttt{[MASK]} token $80\%$ of the time; 
a random token $10\%$ of the time and the unchanged token $10\%$ of the time.
The text is masked in the word level rather than the token level
as in~\cite{KimSK21}.
An MLM head is applied on the outputs of the masked tokens
for prediction with a cross-entropy loss\footnote{Label smoothing is applied in experiments.}, denoted as
$l_{\text{MLM}}$.
%denote the masked language modeling loss.

\begin{figure}[t!]
    \centering
    \begin{tabular}{c@{~~~}c}
    \includegraphics[width=0.47\linewidth]{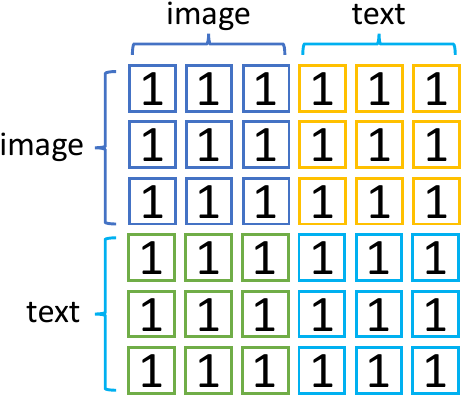} & 
    \includegraphics[width=0.47\linewidth]{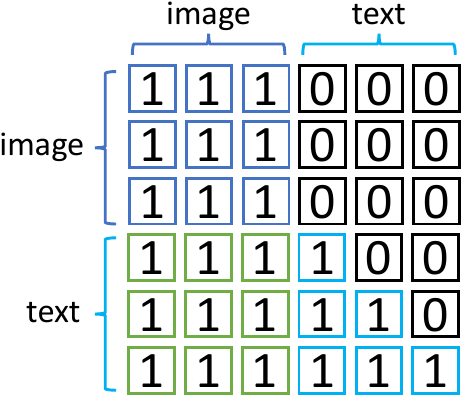} \\ 
    (a) bidirectional & (b) \texttt{seq2seq}
    \end{tabular}
    % \vspace{1mm}
    \caption{Bidirectional and \texttt{seq2seq} attention masks for MLM and S-MLM, respectively.
    If $(i, j)$ is 1, the $i$-th output can depend on the $j$-th input; otherwise, not. 
    }
    \label{fig:mask}
    % \vspace{-3mm}
\end{figure}

When applying the pre-trained model to the image captioning task, we change the attention mask such that the current text token can only depend on the preceding tokens~\cite{Li0LZHZWH0WCG20,abs-2101-00529,abs-2012-06946}. 
Therefore, 
similar as in~\cite{ZhouPZHCG20,abs-1905-03197}, 
we also incorporate another masked language modeling loss 
based on the \texttt{seq2seq} attention, 
and denote it as $l_{\text{S-MLM}}$.
Fig.~\ref{fig:mask} shows the attention mask for MLM and S-MLM.

\begin{table}[t!]
    \centering
    \small
    \begin{tabular}{c@{~~}c@{~~}c@{~~}c@{~~}c}
    \toprule
              & ITC           & ITM             & MLM           & S-MLM \\
    \midrule
    Input     & matched       & (mis)matched    & masked        & masked \\
    % Network   & unimodel      & multimodel      & multimodel    & multimodel \\
    Network Role   & uni.     & multi.     & multi.   & multi.\\
    Attention & bi. & bi.   & bi. & seq. \\
    \bottomrule
    \end{tabular}
    % \vspace{1mm}
    \caption{In different pre-training losses, the input is matched pairs, matched with mismatched pairs, or masked pairs. The network is used as either an unimodal (uni.) encoder or a multimodal (multi.) encoder. The attention mask is either bidirectional (bi.) or \texttt{seq2seq} (seq.).}
    \label{tab:loss_property}
    % \vspace{-2mm}
\end{table}

\subsection{Pre-training Strategies}
\noindent\textbf{One Loss per Iteration}.
Table~\ref{tab:loss_property} summarizes the characteristics of each pre-training loss. 
In different losses, the input format is different, the transformer is used as different roles, and the attention mask may also be different. 
For simplicity, in each iteration, we randomly sample one pre-training loss and calculate the gradient for parameter update. 
% Another way is to calculate the losses for all tasks at each iteration.
% The pro is that the loss is consistent across different iterations and the gradient is stable, but the con is that
% it takes longer training time for each iteration.  
% Empirically, we find the former strategy gives better performance when the total number of loss calculation is the same.
Empirically, we find this strategy is more effective than calculating all losses in each iteration when the total number of loss calculation is the same.

% \vspace{1mm}
\noindent\textbf{Momentum Teacher}.
Motivated by~\cite{abs-2107-07651,abs-1911-05722,abs-2106-09018}, we add a momentum teacher to guide the pre-training. Specifically, the momentum teacher is a clone of the transformer network, and the parameter is updated as the exponential moving average of the target network's parameter. 
Let $\theta$ be the target network's parameter, and $\hat{\theta}$ be the teacher's parameter. Then, in each iteration we have
\begin{equation}
    \hat{\theta} = m \hat{\theta} + (1 - m) \theta,
\end{equation}
where $m$ is set as 0.999 in experiments.
For the pre-training loss of ITC/MLM/S-MLM, the same input and the attention mask are fed to the momentum teacher as well, and the output is used as a soft target of the target network's output. In ITC, let $S_{i, j}=\mathbf{I}_i^T \mathbf{T}_i / t$ be the similarity between the $i$-th image and the $j$-th text, $\hat{S}_{i, j}$ be the similarity from the momentum teacher network. Then, a distillation loss is added on top of ITC as
\begin{align}
    \hat{l}_{\text{ITC}} &= \frac{1}{2}(\hat{l}_1 + \hat{l}_2), \\
    \hat{l}_1  &= \frac{1}{\sum_{i} 1}\sum_{i}\text{KL}(\mathbf{S}_{i, \cdot}, \hat{\mathbf{S}}_{i, \cdot}) \\               
    \hat{l}_2  &= \frac{1}{\sum_{j} 1}\sum_{j}\text{KL}(\mathbf{S}_{\cdot, j}, \hat{\mathbf{S}}_{\cdot, j}),
\end{align}
where $\text{KL}(\cdot, \cdot)$ denotes the Kullback–Leibler divergence loss on the softmax of the inputs.
% the row or column softmax of the $\mathbf{S}$ and $\hat{\mathbf{S}}$.
For MLM loss, let $\mathbf{g}$ be the predicted logits corresponding to the masked token, and $\hat{\mathbf{g}}$ be the logits from the momentum teacher. Then, the distillation loss is $\hat{l}_{\text{MLM}}  = \text{KL}(\mathbf{g}, \hat{\mathbf{g}})$.
% \begin{align}
%     \hat{l}_{\text{MLM}}  = \text{KL}(\mathbf{g}, \hat{\mathbf{g}}).
% \end{align}
Similarly, we have the distillation loss for S-MLM.

\section{Experiments}
\subsection{Experimental Settings}

\paragraph{Network Backbone.}
We use ViT-B/32,
ViT-L/32, and ViT-L/16~\cite{DosovitskiyB0WZ21} as the backbones. 
The number (32 or 16) is the patch size.
ViT-B/32 contains 12 transformer layers with 768 as the hidden size. 
ViT-L/32 and ViT-L/16 have 24 transformer layers with 1024 as the hidden size.
% In pretraining, the backbone is initialized with the imagenet pretrained model.
%as in~\cite{KimSK21}.
Correspondingly, we name our models as UFO-B/32, UFO-L/32 and UFO-L/16. 
% Results over the 3 model are present for comparison with the state-of-the-art and only t
Only the base model (UFO-B/32) is used for ablation studies. 

% \vspace{-2mm}
\paragraph{Vision-Language Pre-training.}
We combine $4$ datasets for vision-language pre-training:
MS COCO~\cite{LinMBHPRDZ14},
% (Karpathy train split~\cite{KarpathyL15}),
Conceptual Captions (CC)~\cite{SoricutDSG18}, 
SBU~\cite{OrdonezKB11}, and Visual Genome (VG)~\cite{KrishnaZGJHKCKL16}.
These datasets result in 4 million images with 10 million associated captions. 
We pre-train the model with $80$ epochs when comparing with the state-of-the-art methods. 
For ablation study, it is 40 epochs unless explicitly specified.
% Each epoch is to scan all image-text pairs once, and in each iteration, only one
% randomly-selected pretraining loss is used for parameter learning as a default setting. 
The image is randomly cropped and resized into $s\times s$ where $s$ ranges from 224 to 384 
with the patch size as the step.
The batch size is $4096$ for UFO-B/32 and UFO-L/32, and $2048$ for UFO-L/16.
% For UFO-B/32, we use $32$ V100 GPUs and $64$ V100 GPUs for UFO-L/32 and UFO-L/16.
The weight decay is $0.01$ and 
is imposed on all parameters except the bias in
linear layers and the layer norm layers. 
The learning rate first linearly warms up to $lr$ and then
linearly decreases to 0.
The learning rate $lr$ is set to $0.0002$ for UFO-B/32 and UFO-L/32, and $0.0001$ for UFO-L/16.
During pre-training, the target model's parameters 
are converted to \texttt{float16} except the loss-related head layers.
The momentum teacher's parameters are kept as \texttt{float32}, 
but the network forward is sped-up by the mixed precision calculation. 
The implementation is based on 
Pytorch\footnote{\url{https://github.com/pytorch/pytorch}}
and Apex\footnote{\url{https://github.com/NVIDIA/apex}}.
Our model is initialized with the ImageNet pre-trained model, and the last checkpoint is used to fine-tune for all downstream tasks.

\begin{table*}[t!]
    \centering
    \small
    \begin{tabular}{@{}c@{}c@{}c@{~~~}c@{~~~}c@{~~~}c@{~~~}c@{~~~}c@{~~~}c@{~~~}c@{~~~}c@{~~~}c@{~~~}c@{~~~}c@{}}
    \toprule
                     & \multirow{2}{*}{Method}                 &  COCO      & \multicolumn{2}{c}{Flickr30k} & \multicolumn{2}{c}{VQAv2}    & COCO        &  \multicolumn{2}{c}{\nocaps} & \multicolumn{2}{c}{NLVR$^2$} & \multicolumn{2}{c}{SNLI-VE}\\
                                                               \cmidrule(r){3-3}     \cmidrule(lr){4-5}  \cmidrule(lr){6-7}              \cmidrule(lr){8-8}  \cmidrule(lr){9-10} \cmidrule(l){11-12} \cmidrule(l){13-14}
                     &                                         &  TR              & TR-ZS            & TR             & test-dev         & test-std          & CIDEr         & CIDEr         &    SPICE        &   dev           & test-P           & dev               & test \\
    \midrule                                                     
\multirow{3}{*}{(a)} & Light.DOT~\cite{sun2021lightningdot}$^D$&  $60.1$          &  -               & $83.9$         & -                &   -               & -             &    -          &     -           &  -              &    -             &   -               & -\\
                     % clip, coco, ZS: 58.4; Align: 58.6                                                                          
                     & CLIP~\cite{RadfordKHRGASAM21}           & -                & $88.0$           & -              & -                &   -               & -             &    -          &     -           &  -              &    -             &   -               & -  \\
                     & ALIGN~\cite{JiaYXCPPLSLD21}             & $77.0$           & $88.6$           & $95.3$         & -                &   -               & -             &    -          &     -           &  -              &    -             &   -               & -  \\
    \midrule                                                                    
\multirow{7}{*}{(b)} & ViLT~\cite{KimSK21}                     &  $62.9$          & $73.2$           & $83.7$         & $71.26$          &   -               &  -            &   -            &    -           &  75.70          & 76.13            &    -              & - \\
                     & SOHO~\cite{abs-2104-03135}              &  $66.4$          & -                & $86.5$         & $73.25$          &  $73.47$          & -             &  -             &    -           &  76.37          & 77.32            &    \textbf{85.0}           & 84.95  \\
                     & OSCAR~\cite{Li0LZHZWH0WCG20}$^{D,S}$    &  $73.5$          & -                &  -             & $73.61$          &  $73.82$          & $127.8$       &  80.9          &    11.7        &  79.12          &  80.37           &    -              & -  \\
                     % in the paper of OSCAR, UNITER's NLVR's dev is 78.40, but in the original paper of UNITER, it is 79.12. Here we respect the original paper. 
                     & UNITER~\cite{ChenLYK0G0020}$^D$         &  $65.7$          & $83.6$           & $87.3$         & $73.82$          &  $74.02$          & -             &   -            &     -          &  79.12          & 79.98            &   79.39           & 79.38 \\
                     & VisParsing~\cite{xue2021probing}        &  -               & -                & $87.0$         & $74.00$          & $74.17$           & -             & -              & -              & 77.61           & 78.05            & 84.75             & \textbf{85.08} \\
                     & VILLA~\cite{abs-2006-06195}$^D$         &   -              & -                & $87.9$         & $74.69$          &  $74.87$          & -             &   -            &     -          &  79.76          & 81.47            &  80.18            & 80.02 \\
                     & ALBEF~\cite{abs-2107-07651}(4M)$^M$     &  $73.1$          & $90.5$           & $94.3$         & $74.54$          &  $74.70$          & -             &   -            &     -          &  80.24          & 80.50            &    -              & -  \\
                     & ALBEF~\cite{abs-2107-07651}(14M)$^M$    &  \textbf{77.6}   & \textbf{94.1}    & \textbf{95.9}  & $75.84$          &  $76.04$          & -             &   -            &     -          &  82.55          & 83.14            &    -              & -  \\
                     & VinVL~\cite{abs-2101-00529}$^{D, S}$    &  $75.4$          & -                & -              & $76.52$          &  $76.60$          & $130.8$       & \textbf{92.46} & 13.07          & \textbf{82.67}  & \textbf{83.98}   &    -              & -  \\
                    %  & SimVLM~\cite{wang2021simvlm} (1.8B)     &  -         & -         & -              & $80.03$  &  $80.34$          & $143.3$       & $112.2$     &    110.3    & 84.53  & 85.15            &  86.21   & 86.32 \\
    \midrule                 
                     
\multirow{3}{*}{(c)} & UFO-B/32(4M)                            &  $74.1$         &  $71.6$           & $91.5$         & $74.21$          & 74.32             & $122.8$       &  78.79         & 12.47          & 76.35           &  76.79           & 77.90             & 77.41   \\ % 10M_BASE_DISTILL_CILP_LEARNABLE_1_0_80E
                     & UFO-L/32(4M)                            &  $76.9$         &  $78.8$           & $93.6$         & 75.73            & $75.74$           & $128.5$       &  88.54         & 13.31          & 78.27           & 78.37            & 78.13             & 77.65    \\% VILT_LARGE32_LEARN_TEMPERATURE_ONLINE_DISTILL_1_0_80E 
                     & UFO-L/16(4M)                            &  $75.7$         &  $74.0$           & $94.1$         & \textbf{76.64}   & \textbf{76.76}    &\textbf{131.2} &  92.26         & \textbf{13.61} & 78.76           & 79.55            & 78.46             & 78.56 \\% VILT_LARGE16_LEARN_TEMPERATURE_ONLINE_DISTILL_1_0_80E_C 
    \bottomrule
    \end{tabular}
    % \vspace{1mm}
    \caption{Compare our model UFO with the state-of-the-art methods. 
    The retrieval task is reported with the top-1 recall 
    on the text retrieval (TR).
    TR-ZS: zero-shot TR. 
    Superscript of $M$: in retrieval tasks, the candidates are filtered by the inner product based on unimodal encoders and then refined by the heavy fusion. 
    Superscript of $D$: the approach depends on an object detector. 
    Superscript of $S$: SCST~\cite{RennieMMRG16} is applied for \nocaps. 
    Rows of (a)/(b)/(c): corresponding to Fig.~\ref{fig:light_heavy} (a)/(b)/(c), respectively. 
    Methods in (b) are slower than those in (a) and (c) as 
    they require to compute the similarity through a heavy-fusion network 
    with each or filtered candidate.
    The number in the parenthesis is the number of images in VLP. \nocaps is on the test set.
    % double checked each number
    }
    \label{tab:compare_sota}
    % \vspace{-2mm}
\end{table*}

\begin{table}[t!]
    \centering
    \small
    \begin{tabular}{ccccc}
   \toprule
    \multirow{2}{*}{Loss}         & VQA  & \multicolumn{2}{c}{Zero-Shot TR@1} \\
         \cmidrule(r){2-2}     \cmidrule{3-4}
         & \texttt{test-dev}  & Flickr30k & COCO \\ 
    \midrule
    % 10M_BASE_FULL_LOSS_SMOOTH_CLIP_WEIGHT1_ALL_LOSS, training hours = 12
    Full &       70.23 & 64.5        & 55.5 \\
    % 10M_BASE_FULL_LOSS_SMOOTH_CLIP_WEIGHT1_TEMPERATURE, 69.1 on flickr; 87.2 for coco tr, training hours = 14
    Random &  71.39 & 68.7        & 58.7\\
        \bottomrule
    \end{tabular}
    % \vspace{1mm}
    \caption{Comparison between the full loss and randomly-selected loss in each iteration. 
    % The epochs are adjusted such that the total number of loss calculation is identical.
    Momentum teacher is disabled.
    }
    \label{tab:single_loss_or_full_loss}
    % \vspace{-3mm}
\end{table}

% \vspace{-2mm}
\paragraph{Downstream Evaluation.}
To evaluate the performance on the light-fusion preferred tasks, we mainly focus on the image-text retrieval task based on the inner product as detailed below. 
For the heavy-fusion favored tasks, we evaluate the performance on VQA, image captioning,
NLVR$^2$, and SNLI-VE. 

% \vspace{1mm}
\noindent\textbf{1) Image-text retrieval.}
The task is to retrieve similar text descriptions based on the image or vice versa. 
The key is to score the similarity between an image and a text description.
As our pre-training incorporates the ITC loss, the similarity can be calculated by the inner product without finetuning for zero-shot (ZS) application.
In the finetuning stage, we simply continue to train the network with the ITC loss. 
The image encoder and the text encoder are not shared
for higher accuracy and initialized 
from the same pre-trained model.
Experiments are conducted on MS COCO~\cite{LinMBHPRDZ14} and 
Flicker30k~\cite{YoungLHH14} datasets with the Karpathy split~\cite{KarpathyL15}.
The top-$K$ recall is reported for the corresponding test set.

% The approach based on heavy fusion is to score the similarity by feeding the concatenation of the image features and the text features into the transformer network for similarity prediction. With $N$ candidates, the approach needs to run the network forward $N$ times, which is time-consuming but may give better accuracy. The approach based on light fusion is to map each image and each description into a global descriptor, and then apply the inner product to score the similarity. As the descriptor can be calculated in the offline, this approach can be much faster. The former approach requires deep understanding of the two input modalities, and the network structure typically follows Fig.~\ref{fig:light_heavy} (b). The latter approach requires less fusion and is normally based on Fig.~\ref{fig:light_heavy} (a).

\noindent\textbf{2) Visual Question Answering (VQA)}. 
The task~\cite{GoyalKSBP16} is to answer a question with natural language based on the image context, and thus requires a deep understanding of the question and the image.
As a common practice, we cast it as a classification problem where each class corresponds to one answer.
The network input is the concatenation of the image and the question embeddings. 
The representation of the text \texttt{[CLS]} is used to predict the answer
over a shared set of $3129$ answers with an MLP layer.
The loss is the binary cross-entropy loss, and the inference is to select the answer with the highest confidence.

\noindent\textbf{3) Image Captioning}.
The task is to describe an image with a
natural language sentence.
As we have S-MLM loss, we reuse this cross-entropy loss to finetune the network on the downstream dataset. Instead of using the word-level masking in
pre-training, we change it to the token-level masking.
In inference, the \texttt{[MASK]} token is appended recursively to the generated tokens to predict the next token
one by one. The beam search size is set as 1, and the accuracy is
evaluated with BLEU@4~\cite{PapineniRWZ02},
METEOR~\cite{DenkowskiL14}, CIDEr~\cite{VedantamZP15}, and SPICE~\cite{AndersonFJG16}.
No SCST~\cite{RennieMMRG16} and CBS~\cite{AndersonFJG16a} are applied.
The dataset is COCO~\cite{LinMBHPRDZ14} with Karpathy
split~\cite{KarpathyL15}, and the model is also evaluated against the val and test sets of the
\nocaps~\cite{abs-1812-08658} benchmark.

% \vspace{1mm}
\noindent\textbf{4) Natural Language Visual Reasoning for Real (NLVR$^2$)}.
The task's input is a pair of images and a natural description, and the goal~\cite{SuhrZZZBA19} is 
to predict whether the description is true about the image pair. 
To fine-tune the network, we construct two input sequences, each containing the
concatenation of the description and one
image. Each sequence is fed to the transformer, and the two outputs
corresponding to \texttt{[CLS]} are concatenated as the joint
representation for a binary linear classifier through
an MLP layer.

% \vspace{1mm}
\noindent\textbf{5) Visual Entailment}.
The task is evaluated on SNLI-VE~\cite{abs-1901-06706} and is to predict the relation between a premise and a sentence hypothesis as one of entailment, neutral or contradiction. 
The premise here is an image for the VL task. 
To finetune the model, we append an MLP layer on top of the text \texttt{[CLS]} token as a three-way classification task. 
The network input is the 
concatenation of the image patch features and hypothesis embeddings.

% \vspace{1mm}
\noindent\textbf{Hyperparameters}.
% For all fine-tuning tasks,
% the input image is resized such that the shorter side is not larger than $S$ and the longer side is not larger than $1333S/800$ while the aspect ratio is kept. 
% The size parameter $S$ is 384 in the ablation study as default 
The input image size is 384 in the ablation study as default
and is increased properly in the comparison with the state-of-the-art methods. 
The batch size is 512 and the model is fine-tuned with $20$ epochs. 
The last checkpoint is used for evaluation.
Other hyperparameters are summarized in the supplementary materials.

\begin{table*}[t!]
    \centering
    \small
    \begin{tabular}{@{}c@{~~}ccccccccccc@{}}
    \toprule
    & \multicolumn{4}{c}{Pre-training Task} & \multicolumn{2}{c}{Zero-Shot Performance} & \multicolumn{5}{c}{Finetune Performance}     \\
    \cmidrule(r){2-5} \cmidrule(lr){6-7} \cmidrule(l){8-12}
         &ITC  & ITM  &   MLM  & S-MLM  & F. TR@1      & C. Caption  & {C. TR@1}     & {F. TR@1}       & VQA            & C. Caption         & \nocaps           \\
\midrule                                   
  (a)    & $\checkmark$   &      &        &        & 54.5         & 0.0         &  65.5         & 83.6            & 68.96          & 108.7              & 55.98           \\ % 10M_BASE_FULL_LOSS_SMOOTH_CLIP_WEIGHT1_TEMPERATURE_CLIP_ONLY 
  (b)    &     &  $\checkmark$   &        &        & 0.1          &  0.0        &   -           & -               & 68.12          & 100.0              & 48.39           \\ % 10M_BASE_FULL_LOSS_SMOOTH_CLIP_WEIGHT1_TEMPERATURE_ITM_ONLY
  (c)    &     &      &   $\checkmark$    &        & 5.0          &  20.3       &  62.5         & 78.4            & 69.84          & 112.1              & 71.46           \\ % 10M_BASE_FULL_LOSS_SMOOTH_CLIP_WEIGHT1_TEMPERATURE_MLM_ONLY  
  (d)    &     &      &        &  $\checkmark$     & 2.3          & \bf84.8       &  61.8         &  78.8           & 69.36          & 115.0              & 75.94           \\ % 10M_BASE_FULL_LOSS_SMOOTH_CLIP_WEIGHT1_TEMPERATURE_SEQ_ONLY 
\midrule                                          
  (e)    & $\checkmark$   &      &   $\checkmark$    &        & \textbf{72.1}& 23.9        &  70.3         & 88.0            & 71.26          &  114.9             & 72.95           \\ % 10M_BASE_FULL_LOSS_SMOOTH_CLIP_WEIGHT1_TEMPERATURE_CLIP_MLM
  (f)    & $\checkmark$   &      &        &   $\checkmark$    & 70.5         & 72.2        &  70.3         & \textbf{89.4}   & 70.30          &  116.5             & 75.32           \\ % 10M_BASE_FULL_LOSS_SMOOTH_CLIP_WEIGHT1_TEMPERATURE_CLIP_SEQMLM
  (g)    &     & $\checkmark$    &    $\checkmark$   &        & 0.1          & 15.0        & 63.2          & 70.8            & 71.33          & 113.1              &  71.42          \\ % 10M_BASE_FULL_LOSS_SMOOTH_CLIP_WEIGHT1_TEMPERATURE_ITM_MLM
\midrule                                                
  (h)    & $\checkmark$   &      &    $\checkmark$   &   $\checkmark$    & 71.0         & 71.7        &  70.0         &  87.0           & 71.31          & 116.8              & 76.46           \\ % 10M_BASE_FULL_LOSS_SMOOTH_CLIP_WEIGHT1_TEMPERATURE_NO_ITM 
  (i)    & $\checkmark$   & $\checkmark$    &        & $\checkmark$      & 68.9         &72.1 &  70.6         & 87.7            & 71.24          & 117.5              & 74.96           \\ % 10M_BASE_FULL_LOSS_SMOOTH_CLIP_WEIGHT1_TEMPERATURE_NO_MLM   
  (j)    & $\checkmark$   & $\checkmark$    &    $\checkmark$   &        & 68.3         & 18.8        & \textbf{71.1} &  88.3           & 71.84          & 116.5              & 73.15           \\ % 10M_BASE_FULL_LOSS_SMOOTH_CLIP_WEIGHT1_TEMPERATURE_NO_SEQ   
  (k)    &     & $\checkmark$    &   $\checkmark$   & $\checkmark$      & 0.2          & 84.5        & 63.6          & 78.2            & 70.86          & 117.8              &  76.35           \\ % SEQ_SMOOTH_DROP_PATH_RECT, parameter is good   
\midrule             
  (l)    & $\checkmark$   & $\checkmark$    &    $\checkmark$   & $\checkmark$      & 70.3         & 70.7        &  70.6         & 88.0            & \textbf{71.87} &  \textbf{119.0}    & \textbf{77.09}    \\ % 10M_BASE_FULL_LOSS_SMOOTH_CLIP_WEIGHT1_TEMPERATURE_80E
    \bottomrule
    \end{tabular}
    % \vspace{1mm}
    \caption{Impact of different pre-training tasks on downstream tasks with 20 epochs for each pre-training loss. 
    C. TR@1: Text Retrieval at top-$1$ on COCO; F. TR@1: Text Retrieval at top-$1$ on Flickr30k. C. Caption: captioning performance in CIDEr on COCO.
    VQA is on \texttt{test-dev}. \nocaps is on val.
    Momentum teacher network is disabled. The highest number for each task is bolded.
    }
    \label{tbl:varying_pre_20epoch_each}
    % \vspace{-2mm}
\end{table*}

\subsection{Comparison with the State-of-the-art}
Table~\ref{tab:compare_sota} shows the comparison with the existing state-of-the-art methods, which are divided into two groups: light-fusion in rows
of (a) and heavy-fusion in rows of (b). Based on the results, we make the following discussions.

% \vspace{1mm}
\noindent\textbf{Applicability on downstream tasks.}
The light-fusion approaches (in rows of (a))
% \eg CLIP~\cite{RadfordKHRGASAM21} and ALIGN~\cite{JiaYXCPPLSLD21},
achieve strong performance on the COCO and Flickr30k retrieval tasks, but are less applicable to other VL tasks such as VQA. 
In contrast, the heavy-fusion approaches (in rows of (b)) are more suitable for the understanding tasks, as multiple transformer layers are dedicated to learn the relationship between the modalities.
Table~\ref{tab:compare_sota} contains the results on retrieval tasks for heavy fusion, 
but the approaches use the fusion network to score the similarity 
between the query and each of the candidates. This process is time-consuming~\cite{sun2021lightningdot}, especially when the dataset is large.
Comparably, our model (in rows of (c)) can achieve competitive retrieval performance with fast speed similar to the light-fusion approaches, and also strong results for other VL understanding tasks.
% is capable to easily adapt to all these tasks as the transformer is pretrained as both the unimodal network and the multimodel fusion network.
% For retrieval tasks, we adopt similar strategies as the ones in the row (a) and 
% can be much faster than the ones in the row (b). Meanwhile, our model can also be well fine-tuned for understanding tasks.

% \vspace{1mm}
\noindent\textbf{Backbones.}
Our UFO-B/32 shares the same backbone with ViLT~\cite{KimSK21}, but achieves significantly stronger performance except on NLVR$^2$, where our approach is slightly better. 
For example of the COCO retrieval task, our model improves ViLT~\cite{KimSK21} from $62.9$ to $72.1$ for the text retrieval at the top-$1$.
On VQA, the improvement is from $71.26$ to $74.21$ on \texttt{test-dev}.
Most methods rely on an object detector to extract object features, 
while our approach removes this dependency and is also in line with other end-to-end approaches. 
Comparing UFO-B/32 with UFO-L/32, we can see a stronger backbone leads to a moderate or significant improvement on all tasks. 

% \vspace{1mm}
\noindent\textbf{Retrieval tasks.}
Our best model (UFO-L/32 or UFO-L/16) achieves better or comparative results than all approaches 
except ALBEF with 14 million images, CLIP with 400 million image-text pairs, and 
ALIGN with 1.8 billion image-text pairs, all of which use a substantially larger pre-training dataset than ours.
Compared to ALBEF~\cite{abs-2107-07651}  with the same amount of images,
our model achieves
better retrieval performance (76.9 vs. 73.1) on COCO and comparable fine-tuned performance on Flickr30k (94.1 vs. 94.3). 
It is worth noting that ALBEF 
refines the retrieval results with a fusion network, while we simply use the inner product for fast retrieval speed.

% \vspace{1mm}
\noindent\textbf{Understanding tasks.}
On the challenging VQA task, our model achieves new state-of-the-art accuracy with 
76.76 on \texttt{test-std}.
This is better than VinVL~\cite{abs-2101-00529} (76.60), which relies on a strong object detector, and 
better than ALBEF~\cite{abs-2107-07651}(14M) (76.04), which uses even more image-text pairs.
On the COCO captioning task, we achieve 131.2 CIDEr score, slightly higher than the previous state of the art (130.8) with cross-entropy optimization.
On \nocaps, our model achieves the best performance in SPICE (13.61 vs. 13.07)
and is competitive in CIDEr (92.26 vs. 92.46).

% except SimVLM~\cite{wang2021simvlm}, which leverages 1.8 billion proprietary image-text pairs. 

% One limitation of our approach is the inferior performance on NLVR2 and SNLI. We suspect that the two tasks may require more understanding capability

% \noindent\textbf{Patch size}
% Comparing UFO-L/16 with UFO-L/32, we can observe a smaller patch size leads to better performance in understanding tasks and comparable results in retrieval tasks. 

\subsection{Ablation Study}

\paragraph{Different Pre-training Strategies.}
We randomly choose one task in each iteration.
An alternative is to run all tasks in each iteration where the gradient is more stable.
To make a fair comparison, we adjust the training epochs such that the total number of loss calculations is the same,
in which the training cost is roughly the same. 
The comparison is shown in Table~\ref{tab:single_loss_or_full_loss}, and we can see that the accuracy with the randomly-selected loss shows better accuracy than full losses in each iteration. Therefore, the model may favor more iteration updates than more stable gradients.

\begin{table*}[t!]
    \centering
    \small
    \begin{tabular}{cccccccccc}
    \toprule
        & \multicolumn{2}{c}{w/ or w/o Momentum Teacher} & \multicolumn{7}{c}{Downstream Performance}              \\
    \cmidrule(r){2-3}\cmidrule(l){4-10}
        & Pre-training     & Finetuning     & C. TR@1 & F.TR@1 & VQA      & C. Caption  & \nocaps & NLVR$^2$ & SNLI-VE     \\
    \midrule 
    (a) &                 &                & 70.6    &  88.0  & 71.87    & 119.0    & 77.09  & 75.4 & 77.4         \\ % 10M_BASE_FULL_LOSS_SMOOTH_CLIP_WEIGHT1_TEMPERATURE_80E
    (b) &                 & $\checkmark$   & 71.3    & 88.4   & 71.94    & 118.6    & 76.76  & 75.6 & 76.9         \\
    (c) & $\checkmark$    &                & 72.1    &  89.9  & 72.42    & 119.4    & 77.53  & 76.2 & 77.5         \\ % 10M_BASE_DISTILL_CILP_LEARNABLE_1_0_80E
    (d) & $\checkmark$    & $\checkmark$   & 72.0    & 89.2   & 72.51    & 119.9    & 78.03  & 76.2 & 77.8         \\
    % (d)
    % vqa ema, weight; (0.01, 71.92); (0.001, 72.46); (0.0002, 72.51); (0.0001, 72.42), (2e-5, 72.48), (1e-5, 72.45)
    % caption; (0.001, 119.5), (0.01, 119.8), (0.1, 119.9), (0.2, 119.9), (1.0, 119.4)
    % nocaps: (0.001, 77.85), (0.01, 77.66), (0.1, 77.36), (0.2, 78.03), (1.0, 77.14)
    % c. tr@1: (0.001, 71.9), (0.01, 72.0), (0.1, 71.8), (0.5, 71.6), (1.0, 71.3)
    % F. TR@1: (0.001, 88.7), (0.01, 88.5), (0.1, 88.8), (1.0, 89.2)
    \bottomrule
    \end{tabular}
    % \vspace{1mm}
    \caption{Effectiveness of the momentum teacher in pre-training and finetuning stages. VQA is on \texttt{test-dev}. \nocaps, NLVR$^2$ and SNLI-VE are on the validation split.
    Pre-training is with 80 epochs.}
    \label{tab:momentum_teacher}
    % \vspace{-2mm}
\end{table*}

\begin{table*}[t!]
    \centering
    \small
    \begin{tabular}{cccccccc}
    \toprule
    Input size                  &  VQA           & COCO Caption     & NOCAPS          & Flickr-TR@1    & COCO-TR@1      & NLVR2          & SNLI-VE        \\
    \midrule                    
    % 10M_BASE_DISTILL_CILP_LEARNABLE_1_0_80E                   
    384                         & 72.42          & 119.4            & 77.53           & 88.8           & 72.1           & 76.2           &  77.5               \\
    \midrule                                        
    480                         & 73.29          & 121.4            & 79.21           & 88.3           & 72.7           & \textbf{76.4}  &  77.6               \\
    576                         & 72.87          & 121.8            & 79.72           & 89.4           & 72.3           & 76.2           &  \textbf{77.9}               \\
    672                         & 74.00          & 122.0            & 80.00           & 90.8           & 72.8           & 75.7           &  77.8               \\
    768                         & \textbf{74.21} & \textbf{122.8}   & \textbf{80.74}  & 90.8           & 73.6           &  -             &  77.7               \\
    960                         & 73.49          & 122.8            & 80.01           & 91.3           & \textbf{74.1}  &  -             &  77.9               \\
    1024                        & 73.07          &  -               &  -              & \textbf{91.5}  & 73.9           &  -             &  -               \\
    1280                        & 73.49          &  -               &  -              & 90.5           & 74.0           &  -             &   -                 \\
    \bottomrule
    \end{tabular}
    % \vspace{1mm}
    \caption{Impact of different image sizes during finetuning. VQA is on \texttt{test-dev}; \nocaps, NLVR$^2$ and SNLI-VE are on val. 
    % number is double verified.
    }
    \label{tab:vary_input}
    % \vspace{-2mm}
\end{table*}

% \vspace{-2mm}
\paragraph{Different Pre-training Tasks.}
We have multiple pre-training losses, and 
one question is that how each loss impacts 
the performance of the downstream tasks.
We conduct experiments in two settings.
The first is to run each loss $20$ epochs to
study whether more losses help the downstream tasks, and results are shown in Table~\ref{tbl:varying_pre_20epoch_each}. 
On VQA, COCO Caption and \nocaps, 
the one with all pre-training tasks leads to the highest accuracy after fine-tuning.
For the other tasks, 
we can also observe competitive accuracy with
all losses.
Other observations are detailed below.
\begin{enumerate}[leftmargin=0.5cm]
    \item (a) vs. (b) vs. (c) vs. (d): With a single pre-training loss, S-MLM and ITC give the best performance on captioning and retrieval tasks, respectively. This is reasonable as the pre-training task is consistent with the corresponding downstream task. 
    For VQA task, MLM gives the best accuracy.
    \item (a) vs. (e) or (f): MLM or S-MLM improves the ITC task and shows significant improvement on the retrieval task, \eg, from $83.6\%$ to $88.0\%$ for the text retrieval at the top-$1$ with finetuning on Flickr30k.
    With MLM or S-MLM, the captioning task and VQA can also be 
    improved by a large margin on top of the ITC loss.
    \item (c) vs. (e); (d) vs. (f): 
    On top of MLM or S-MLM, ITC gives clear improvement for VQA, \eg, 69.36 (b) to 70.30 (e).
    \item (l) vs. (h, i, j, k): For VQA, we can see significant accuracy drop by removing ITM, or MLM, or ITC, compared with all pre-training losses.
    In spite of less drop (71.87 $\rightarrow$ 71.84) on VQA by removing S-MLM, the captioning task drops a lot (77.09 to 73.15 on \nocaps). This also shows all these pre-training losses help on the VL understanding tasks. 
    For retrieval tasks, the accuracy is similar as long as ITC and at least MLM or S-MLM exists.
\end{enumerate}
The second experiment setting is that the total number of pre-training epochs is $40$ such that
the pre-training cost is roughly the same as only one loss is randomly selected in each iteration.
Overall, we observe strong performance with all pre-training losses. The experiment details are in the supplementary materials.

\paragraph{Momentum Distillation.}
We apply the momentum distillation~\cite{abs-2107-07651} to regularize the model training in the pre-training stage.
Table~\ref{tab:momentum_teacher} shows the ablation study by turning it on or off in both pre-training and finetuning stages. 
Comparing (a) with (c), we can see the performance improves significantly on VQA and retrieval tasks by enabling the momentum distillation in the pre-training stage,
and shows slight improvement on other tasks.
However, during the fine-tuning stage\footnote{We search the best weight among $\{1.0, 0.1, 0.01, 0.001\}$
on the distillation loss in the fine-tuning. A finer hyperparameter search may potentially improve the results}, 
it shows almost little improvement in our setting.
The reason may be that the datasets in the downstream tasks are well annotated, 
while the massive pre-training dataset is noisy.
As the momentum teacher can reduce
the impact of the data noise~\cite{abs-2107-07651},
it helps more the pretraining than the fine-tuning.

\paragraph{Multi-scale vs. single-scale.}
In VLP, we use the multi-scale 
image inputs ($224\sim 384$).
Another way is to use the single-scale input ($384$ always). 
The former could reduce the training time as the size is smaller or equal to $384$, and the model can be more robust for scale changes.
Table~\ref{tab:multi_scale} shows the comparison.
On VQA, the accuracy with the multi-scale improves, while in the zero-shot retrieval task on Flickr30k, the performance drops.
Considering the reduced training cost, we always choose the multi-scale.

\begin{table}[t!]
    \small
    \centering
    % \begin{tabular}{@{}c@{~~}c@{~~}c@{~~}c@{}}
    \begin{tabular}{cccc}
    \toprule
    Scale            & Hours &  VQA      & ZS Flickr TR@1  \\ % & 0-Coco-TR \\
    \midrule
    % 10M_BASE_SINGLE_SCALE_CLIP1_LEARNT, 17 hours, 1 minutes
    single           & 17             & 71.19     & 70.4          \\ %  & 53.6           \\
    % 10M_BASE_FULL_LOSS_SMOOTH_CLIP_WEIGHT1_TEMPERATURE, 14 hours + 20 minutes
    multi            & 14             & 71.39     &  68.7         \\ %  & 58.7       \\
    \bottomrule
    \end{tabular}
    \caption{Comparison between the multi-scale and  single-scale image inputs during VLP on 32 V100. }
    \label{tab:multi_scale}
\end{table}

% \vspace{-2mm}
\paragraph{Increasing the image size}
Table~\ref{tab:vary_input} shows the study with different input image sizes for each
downstream task.
On VQA, image captioning and retrieval tasks, the accuracy improves a lot with a larger input size. 
For example on VQA, the accuracy is improved from $72.42$ to $74.21$ by increasing the input size from 384 to 768. 
On NLVR$^2$ and SNLI-VE, the improvement is minor.
Meanwhile, the optimal input size is also different on different tasks, which indicates that different tasks may expect different granularity levels of image understanding.

\section{Conclusion}
We propose a single unified transformer (UFO), which is capable of processing both the unimodal input of images or text, and the multimodal input. 
During the vision-language pre-training, the network is learned to understand different signals through multiple losses, including the image-text contrastive loss, the image-text matching loss, and the masked language modeling loss based on the bidirectional and \texttt{seq2seq} masks.
Extensive experiments show that our unified model can achieve competitive results compared to existing methods, which typically design specific networks 
for each modality and modality fusion.
As our model is up to large-sized (24 layers) on only 4 million images in VLP, 
we expect it is beneficial to scale up 
both the model size and the pre-training data,
which is left as future work.

\appendix
\newpage

\begin{table}[]
    \centering
    \small
    \begin{tabular}{c@{~}c@{~}c@{~}c@{~}c@{~}c}
    \toprule
     Datasets     & COCO~\cite{LinMBHPRDZ14}    & CC~\cite{SoricutDSG18}   & SBU~\cite{OrdonezKB11}    & VG~\cite{KrishnaZGJHKCKL16}    & Total \\
     \midrule
    \#Images      & 113K    & 3.1M & 875K   & 108K  & 4.2M \\
    \#Captions    & 567K    & 3.1M & 875K   & 5.4M  & 9.9M \\
    \bottomrule
    \end{tabular}
    % CC: 3101621; SBU: 874806; both of which has 1 cpation for each image
    % vg: 5407172; 108045
    \caption{Dataset statistics in VL pretraining. }
    \label{tbl:data_statistics}
\end{table}

\begin{table}[]
    \centering
    \small
    \begin{tabular}{@{}c@{~}c@{~~}c@{~~}c@{}}
    \toprule
    Model              & UFO-B/32 & UFO-L/32 & UFO-L/16 \\
    \midrule
    Batch size         & 4096     & 4096     & 2048     \\
    Number of V100     & 32       & 64       & 64       \\
    \midrule
    Max GPU Mem (GB)   & 17       & 16       & 27        \\
    Time cost (hour)   & 32       & 60       & 177       \\
    \bottomrule
    \end{tabular}
    \caption{Vision language pretraining statistics for 80 epochs in our experiment. 
    With different implementations and hardware settings,
    the cost can be greatly different.
    }
    \label{tab:vlp_cost}
\end{table}

The supplementary materials follow the same structure with the main paper, but provides more details and studies.

\section{Experiment}
\subsection{Settings}
\subsubsection{Vision-Language Pre-training}
\noindent\textbf{Dataset.}
Table~\ref{tbl:data_statistics} shows the statistics of each dataset. 

\noindent\textbf{Data Preprocessing.}
For faster data loading, we downsize each image such that the shorter side is not larger than $384$ and the longer side is not larger than $640$ while keeping the aspect ratio. 
Meanwhile, all images are re-compressed as the JPEG format with the quality being $90$.
In Visual Genome, as the caption is region-level, for each caption we crop a sub region which is $4$ times of the associated box annotation, and take this extended region as the input image.

\noindent\textbf{Data Augmentation.}
We apply the multi-scale and random cropping as the data augmentation. 
Specifically, the cropped region area takes at least $80\%$ of the whole image\footnote{Implemented as RandomResizedCrop(scale=(0.8, 1.0),ratio=(1.0, 1.0)) in Pytorch}. The region is resized into $s\times s$ and $s$ is a random value ranging from $224$ to $384$ with the patch size as the step. 

\noindent\textbf{Pretraining Statistics}. 
Table~\ref{tab:vlp_cost} shows the vision langauge pretraining statistics with $80$ epochs. 

\subsubsection{Downstream Evaluation}
During fine-tuning, no preprocessing, \eg JPEG re-comopression, is applied.
The image is resized such that the shorter side is not larger than $s$ 
and the longer 
side is not longer than $1333s/800$ with the aspect ratio kept. 
No random crop and multi-scale augmentation are applied here.
Following~\cite{KimSK21}, we apply the RandAugment~\cite{abs-1909-13719} except the color inversion to the image.
Table~\ref{tab:input_size_lr} shows the input size $s$ and the learning rate 
when comparing with the state-of-the-art approaches (shown in Table 2 of the main paper).
The corresponding learning rate is the default setting for the ablation study, where the input size $s$ is $384$ as default.

\begin{table*}
    \centering
    \small
    \begin{tabular}{@{}ccccccccc@{}}
                \toprule
\multirow{2}{*}{Model}    & \multirow{2}{*}{Param}              & \multicolumn{2}{c}{Retrieval} & \multirow{2}{*}{VQA} & \multicolumn{2}{c}{Captioning} & \multirow{2}{*}{NLVR2} & \multirow{2}{*}{SNLI-VE} \\
                                                                \cmidrule(lr){3-4}                                     \cmidrule(lr){6-7}
                          &                                     & COCO      & Flickr            &                      & COCO          & NOCAPS         &                        &                          \\
                          \midrule                      
\multirow{2}{*}{UFO-B/32} & Input size                          &  960      &  1024             &  768                 & 768           & 768            &  480                   &  576                    \\
                          & Learning rate                       &  2.5e-5   & 2.5e-5            &  5e-5                & 5e-5          & 5e-5           &  5e-5                  &  2.5e-5             \\ % 10M_BASE_DISTILL_CILP_LEARNABLE_1_0_80E
                          \midrule                      
\multirow{2}{*}{UFO-L/32}  & Input size                         &  768      & 768               &  768                 & 768           & 768            & 384                    & 480                            \\
                          & Learning rate                       &  1e-5     & 1e-5              &  5e-5                & 5e-5          & 5e-5           & 5e-5                   & 1e-5                             \\ %VILT_LARGE32_LEARN_TEMPERATURE_ONLINE_DISTILL_1_0_80E
                          \midrule                      
\multirow{2}{*}{UFO-L/16}  & Input size                          &  480     &  384              & 576                  & 576           & 672            &  384                   &  672                        \\ %VILT_LARGE16_LEARN_TEMPERATURE_ONLINE_DISTILL_1_0_80E_C
                          & Learning rate                       &  1e-5     &  1.25e-5          & 3e-5                 & 1.5e-5        & 1e-5           &  5e-5                  &  1e-5                        \\
                          \bottomrule
    \end{tabular}
    \caption{Input size and the peak learning rate in each downstream task when comparing with the state-of-the-art approaches in Table 2 of the main paper. The learning rate is also used as the default setting for all ablation studies.
    NOCAPS shares the same training data with the COCO captioning task.
    % double checked each number
    }
    \label{tab:input_size_lr}
\end{table*}

\subsection{Comparison with the State-of-the-art Approaches}
In the Table 3 of the main paper, 
we present the comparison with the state-of-the-art approaches. 
For the retrieval task, the result is based on the text retrieval at the top-$1$ and for the caption task, 
it is the CIDEr score.
Here, we show the results with other widely-used metrics.
Table~\ref{tab:retrieval} illustrates the complete result on the retrieval task. 
The observation is consistent with the discussion in the main paper. 
Under the same pretraining data scale with ALBEF~\cite{abs-2107-07651}, our approach achieves better accuracy on the COCO dataset and competitive results on Flickr30k. 
Meanwhile, our best model is also competitive compared with the best model with much larger pretraining data scale.
Table~\ref{tab:coco_caption} and~\ref{tab:nocaps} show the complete results on the COCO captioning task and \nocaps dataset, respectively.

\begin{table*}[]
    \centering
    \small
    \begin{tabular}{@{}l@{~~}c@{~~}c@{~~}c@{~~}c@{~~}c@{~~}c@{~~}c@{~~}c@{~~}c@{~~}c@{~~}c@{~~}c@{~~}c@{}}
    \toprule
                     & \multirow{2}{*}{Method}                   &  \multicolumn{6}{c}{Flickr}                        & \multicolumn{6}{c}{COCO}  \\
                                                                  \cmidrule(lr){3-8}                                  \cmidrule(lr){9-14}
                     &                                           & TR@1       &  TR@5   & TR@10     & IR@1    & IR@5    & IR@10   & TR@1    & TR@5    & TR@10   & IR@1    & IR@5    & IR@10 \\
    \midrule                              
\multirow{2}{*}{(a)} & Light.DOT~\cite{sun2021lightningdot}$^D$  &  $83.9$    &  $97.2$ & $98.6$    & 69.9    & 91.1    & 95.2    & 60.1    & 85.1    & 91.8    & 45.8    & 74.6    & 83.8 \\
                     & Align~\cite{JiaYXCPPLSLD21}(1.8B)         & $95.3$     & $99.8$  & $100.0$   & 84.9    & 97.4    & 98.6    & 77.0    & 93.5    & 96.9    & 59.9    & 83.3    & 89.8 \\
    \midrule                                                
\multirow{7}{*}{(b)} & VILT~\cite{KimSK21}                       &  83.5      & 96.7    & 98.6      & 64.4    & 88.7    & 93.8    & 61.5    & 86.3    & 92.7    & 42.7    & 72.9    & 83.1 \\
                & Light.DOT~\cite{sun2021lightningdot}$^{D, M}$  & 87.2       & 98.3    & 99.0      & 75.6    & 94.0    & 96.5    & 74.2    & 92.4    & 96.0    & 57.4    & 82.7    & 89.9 \\
                     & UNITER~\cite{ChenLYK0G0020}$^D$           & 87.3       & 98.0    & 99.2      & 75.6    & 94.1    & 96.8    & 65.7    & 88.6    & 93.8    & 52.9    & 79.9    & 88.0 \\
                     & VILLA~\cite{abs-2006-06195}$^D$           & 87.9       & 97.5    & 98.8      & 76.3    & 94.2    & 96.8    &  -      &   -     &  -      &   -     &  -      &  -   \\
                     & OSCAR~\cite{Li0LZHZWH0WCG20}$^D$          &  -         &  -      &  -        &  -      &  -      &  -      & 73.5    & 92.2    & 96.0    & 57.5    & 82.8    & 89.8 \\
                     & ALBEF~\cite{abs-2107-07651}(4M)$^{M}$     & \bf94.3    & 99.4    & 99.8      & 82.8    & 96.7    & 98.4    & 73.1    & 91.4    & 96.0    & 56.8    & 81.5    & 89.2 \\
                     & VinVL~\cite{abs-2101-00529}$^D$           &  -         &  -      &   -       &  -      &  -      &   -     & 75.4    & 92.9    & 96.3    & 58.8    & 83.5    & 90.3 \\
                     & ALBEF~\cite{abs-2107-07651}(14M)$^{M}$    & 95.9       &\bf99.8  &\bf100.0   & \bf85.6 & \bf97.5 & \bf98.9 & \bf77.6 & \bf94.3 & \bf97.2 & \bf60.7 & \bf84.3 & \bf90.5 \\
    \midrule
    
 \multirow{3}{*}{(c)} & UFO-B/32(4M)                             &  91.5      & 99.2    & 99.7      & 79.0    & 95.2    & 97.6    & 74.1    & 93.2    & 96.8    & 56.4    & 82.4    & 89.5 \\ % 10M_BASE_DISTILL_CILP_LEARNABLE_1_0_80E
                      & UFO-L/32(4M)                             &  93.6      & 99.0    & 99.8      & 81.1    & 95.9    & 98.1    & 76.9    & 94.1    & 97.1    & 59.1    & 83.7    & 90.4 \\% VILT_LARGE32_LEARN_TEMPERATURE_ONLINE_DISTILL_1_0_80E 
                      & UFO-L/16(4M)                             &  94.1      & 99.5    & 99.9      & 80.7    & 96.7    & 98.3    & 75.7    & 93.7    & 97.1    & 59.2    & 83.6    & \bf90.5 \\% VILT_LARGE16_LEARN_TEMPERATURE_ONLINE_DISTILL_1_0_80E_C 
     \bottomrule
    \end{tabular}
    \caption{
    Compare our model UFO with the state-of-the-art approaches in the retrieval task after fine-tuning. 
    Superscript of $M$: the retrieval candidate is first filtered by the innner product and then refined by the heavy fusion.
    Superscript of $D$: the approach is based on an object detector. 
    The number in parenthesis: the number of images in pretraining.
    TR: text retrieval.
    IR: image retrieval.
    The approaches in the rows (a) and (c) are based on the inner product and thus the retrieval speed can be fast.
    % all numbers are double verified. 
    }
    \label{tab:retrieval}
\end{table*}

\begin{table}
    \centering
    \small
    \begin{tabular}{c@{~}c@{~}c@{~}c@{~}c}
    \toprule
 Method                                     & BLEU@4            &  METEOR          & CIDEr          & SPICE \\
                            \midrule            
 MiniVLM~\cite{abs-2012-06946}              & $35.6$            & $28.6$           & $119.8$        & $21.6$ \\
OSCAR~\cite{Li0LZHZWH0WCG20}                & 37.4              & \textbf{30.7}    & 127.8          & \textbf{23.5}  \\
VinVL~\cite{abs-2101-00529}                 & 38.5              & 30.4             & 130.8          & 23.4 \\
% SimVLM(1.8B)~\cite{wang2021simvlm}          & 40.6       & 33.7      & 143.3     & 25.4  \\
    \midrule    
UFO-B/32(4M)                                & 36.0              & 28.9             & 122.8          & 22.2 \\ % 10M_BASE_DISTILL_CILP_LEARNABLE_1_0_80E
UFO-L/32(4M)                                & 37.6              & 29.7             & 128.5          & 23.0 \\% VILT_LARGE32_LEARN_TEMPERATURE_ONLINE_DISTILL_1_0_80E 
UFO-L/16(4M)                                & \textbf{38.7}     & 30.0             & \textbf{131.2} & 23.3 \\% VILT_LARGE16_LEARN_TEMPERATURE_ONLINE_DISTILL_1_0_80E_C 
     \bottomrule
    \end{tabular}
    \caption{
    Compare our model UFO with the state-of-the-art approaches in the captioning task on COCO based on the cross-entropy loss. 
    }
    \label{tab:coco_caption}
\end{table}

\begin{table*}[]
    \centering
    \small
    \begin{tabular}{@{}c@{}c@{~}c@{~~}c@{}c@{}c@{}c@{}c@{~~}c@{~~~~}c@{}c@{}c@{}c@{}c@{}c@{}c@{~~}c@{}}
    \toprule
    \multirow{3}{*}{Method}      & \multicolumn{8}{c}{Validataion set}  & \multicolumn{8}{c}{Test set} \\
                                 \cmidrule(lr){2-9} \cmidrule(l){10-17}
                                 & \multicolumn{2}{c}{{in-domain}}  & \multicolumn{2}{c}{{ne.-domain}}  & \multicolumn{2}{c}{{ou.-domain}} & \multicolumn{2}{c}{{overall}} & \multicolumn{2}{c}{{in-domain}}  & \multicolumn{2}{c}{{ne.-domain}}  & \multicolumn{2}{c}{{ou.-domain}} & \multicolumn{2}{c}{{overall}}  \\
                                 \cmidrule(lr){2-3}               \cmidrule(lr){4-5}                \cmidrule(lr){6-7}                  \cmidrule(lr){8-9} \cmidrule(lr){10-11}
                                 \cmidrule(lr){12-13} \cmidrule(lr){14-15} \cmidrule(l){16-17}
                                         & C              & S       & C          & S           & C          & S         & C          & S        & C          & S        & C            & S           & C          & S         & C          & S         \\
                                           \midrule                         
OSCAR~\cite{Li0LZHZWH0WCG20}$^{S,C}$     & 85.4           & 11.9    & 84.0       & 11.7        & 80.3       & 10.0      & 83.4       & 11.4     & 84.8       & 12.1     & 82.1     & 11.5   & 73.8      & 9.7       & 80.9      & 11.3       \\
    Human~\cite{abs-1812-08658}          & 84.4           & 14.3    & 85.0       & 14.3        & 95.7       & 14.0      & 87.1       & 14.2     & 80.6       & 15.0     & 84.6     & 14.7   & 91.6      & 14.2      & 85.3      & 14.6        \\
    VIVO~\cite{abs-2009-13682}$^{S,C}$   & 92.2           & 12.9    & 87.8       & 12.6        & \bf87.5    & 11.5      & 88.3       & 12.4     & 89.0       & 12.9     & 87.8     & 12.6   & \bf80.1   & 11.1      & 86.6      & 12.4        \\
VinVL~\cite{abs-2101-00529}$^{S}$        & 103.7          & 13.7    & \bf95.6    & 13.4        & 83.8       & 11.9      & \bf94.3    & 13.1     & 98.0       & 13.6     & \bf95.2  & 13.4   & 78.0      & 11.5      & \bf92.5   & 13.1        \\
% SimVLM(1.8B)~\cite{wang2021simvlm} & \textbf{113.7} & -             & \textbf{110.9} & -               & \textbf{115.2} & -             & \textbf{115.2} & -             \\
    \midrule      
    % UFO-B/32                     & 94.52         & 13.44          & 82.65          & 12.80           & 64.86          & 10.97         & 80.74          & 12.54 & 90.64 & 13.56 & 81.31 & 12.71 & 60.64 & 10.65 & 78.79 & 12.47         \\    % 10M_BASE_DISTILL_CILP_LEARNABLE_1_0_80E
    % UFO-L/32                     & 99.79 & 14.02 & 91.12 & 13.36 & 75.59 & 11.74 & 89.22 & 13.14 & 96.05 & 14.11 & 90.87 & 13.53 & 74.17 & 11.84 & 88.54 & 13.31  \\ % VILT_LARGE32_LEARN_TEMPERATURE_ONLINE_DISTILL_1_0_80E
    % UFO-L/16                     & 103.91 & 14.52 & 95.53 & 13.76 & 83.52 & 12.31 & 94.3 & 13.59 & 98.93 & 14.3 & 94.73 & 13.86 & 77.93 & 12.06 & 92.26 & 13.61 \\ % VILT_LARGE16_LEARN_TEMPERATURE_ONLINE_DISTILL_1_0_80E_C
    UFO-B/32                             & 94.5           & 13.4    & 82.7       & 12.8        & 64.9       & 11.0      & 80.7      & 12.5      & 90.6       & 13.6     & 81.3     & 12.7   & 60.6      & 10.7      & 78.8      & 12.5         \\    % 10M_BASE_DISTILL_CILP_LEARNABLE_1_0_80E
    UFO-L/32                             & 99.8           & 14.0    & 91.1       & 13.4        & 75.6       & 11.7      & 89.2      & 13.1      & 96.1       & 14.1     & 90.9     & 13.5   & 74.2      & 11.8      & 88.5      & 13.3  \\ % VILT_LARGE32_LEARN_TEMPERATURE_ONLINE_DISTILL_1_0_80E
    UFO-L/16                             & \bf103.9       & \bf14.5 & 95.5       & \bf13.8     & 83.5       & \bf{12.3} & \bf{94.3} & \bf{13.6} & \bf{98.9}  & \bf14.3  & 94.7     & \bf13.9& 77.9      & \bf12.1   & 92.3      & \bf13.6 \\ % VILT_LARGE16_LEARN_TEMPERATURE_ONLINE_DISTILL_1_0_80E_C
    \bottomrule
    \end{tabular}
    \caption{Compare our model UFO with the state-of-the-art approaches in \nocaps with finetuning. C: CIDEr. S: SPICE. ne.-domain: near-domain. ou.-domain: out-of-domain. 
    The highest score is bolded (The human performance is excluded). Superscript of S: SCST~\cite{RennieMMRG16} is applied. Superscipt of C: CBS~\cite{AndersonFJG16a} is applied.
    % number is verified against the vinvl paper.
    }
    \label{tab:nocaps}
\end{table*}

\subsection{Ablation Study}

\noindent\textbf{Different Pretraining Tasks.}
In the main paper, we studied the results with different pretraining tasks when each task runs with the same number of iterations. 
Table~\ref{tbl:varying_pre_40epoch_total} shows the results when the total number of iterations is the same, and thus the pretraining cost is similar. 
We can see that the pretraining with all losses can achieve descent performance on all downstream tasks compared with the best performance. 
It is noted that the row (j) gives the best VQA performance by removing S-MLM loss, but sacrifice a lot on the captioning tasks. 
Thus, we stick to apply all losses during pretraining.

\begin{table*}[]
    \centering
    \small
    \begin{tabular}{@{}c@{~~}ccccccccccc@{}}
    \toprule
    & \multicolumn{4}{c}{Pretraining task} & \multicolumn{2}{c}{Zero-Shot Performance} & \multicolumn{5}{c}{Finetune Performance}     \\
    \cmidrule(r){2-5} \cmidrule(lr){6-7} \cmidrule(l){8-12}
         &ITC  & ITM &   MLM  & S-MLM  & F. TR@1      & C. Caption   & {C. TR@1}    & {F. TR@1}     & VQA            & C. Caption    & NOCAPS         \\
\midrule                       
 (d)     & \checkmark   &     &        &        & 65.5         & 0.0          &68.3          & 85.5          & 69.58          & 109.1         & 56.91          \\ % 10M_BASE_FULL_LOSS_SMOOTH_CLIP_WEIGHT1_TEMPERATURE_CLIP_ONLY
 (a)     &     & \checkmark   &        &        & 0.2          &  0.0         & -            & -             & 68.37          & 99.0          & 45.86          \\ % 10M_BASE_FULL_LOSS_SMOOTH_CLIP_WEIGHT1_TEMPERATURE_ITM_ONLY
 (b)     &     &     &   \checkmark    &        & 5.0          &  21.9        &62.8          & 78.3          & 70.42          & 113.2         & 71.46          \\ % 10M_BASE_FULL_LOSS_SMOOTH_CLIP_WEIGHT1_TEMPERATURE_MLM_ONLY
 (c)     &     &     &        &  \checkmark     & 0.8          &  78.6        &61.6          &  77.9         & 69.38          & 117.0         & \textbf{77.06} \\ % 10M_BASE_FULL_LOSS_SMOOTH_CLIP_WEIGHT1_TEMPERATURE_SEQ_ONLY
\midrule                              
 (e)     & \checkmark   &     &   \checkmark    &        & \textbf{72.1}& 23.9         &\textbf{70.3} & 88.0          & 71.26          &  114.9        & 72.95          \\ % 10M_BASE_FULL_LOSS_SMOOTH_CLIP_WEIGHT1_TEMPERATURE_CLIP_MLM
 (f)     & \checkmark   &     &        &   \checkmark    & 70.5         & 72.2         &\textbf{70.3} & \textbf{89.4} & 70.30          &  116.5        & 75.32          \\% 10M_BASE_FULL_LOSS_SMOOTH_CLIP_WEIGHT1_TEMPERATURE_CLIP_SEQMLM
 (g)     &     & \checkmark   &    \checkmark   &        & 0.1          & 15.0         & 63.2         & 70.8          & 71.33          & 113.1         &  71.42         \\ % 10M_BASE_FULL_LOSS_SMOOTH_CLIP_WEIGHT1_TEMPERATURE_ITM_MLM
\midrule                             
 (h)     & \checkmark   &     &    \checkmark   &   \checkmark    & 69.0         & 72.2         &  69.8        &  86.0         & 70.93          & 116.1         & 74.65          \\ % 10M_BASE_FULL_LOSS_SMOOTH_CLIP_WEIGHT1_TEMPERATURE_NO_ITM
 (i)     & \checkmark   & \checkmark   &        &   \checkmark    & 66.3         & 72.7         & 70.2         & 87.1          & 71.11          & \textbf{117.1}& 73.64          \\ % 10M_BASE_FULL_LOSS_SMOOTH_CLIP_WEIGHT1_TEMPERATURE_NO_MLM
 (j)     & \checkmark   & \checkmark   &    \checkmark   &        & 68.7         & 25.4         & 69.8         &  87.0         & \textbf{71.65} & 114.7         & 71.26           \\ % 10M_BASE_FULL_LOSS_SMOOTH_CLIP_WEIGHT1_TEMPERATURE_NO_SEQ
 (k)     &     & \checkmark   &    \checkmark   & \checkmark      & 0.1          & \textbf{76.3}& 63.6         & 78.2          & 70.77          & 117.0         & 74.31           \\ % SEQ_SMOOTH_DROP_PATH_RECT, parameter is good
\midrule                            
 (l)     & \checkmark   & \checkmark   &    \checkmark   & \checkmark      & 68.7         & 72.7         &  69.1        & 87.6          & 71.39          &  116.7        & 74.77           \\ % 10M_BASE_FULL_LOSS_SMOOTH_CLIP_WEIGHT1_TEMPERATURE
    \bottomrule
    \end{tabular}
    \caption{Impact of different pretraining tasks on downstream tasks with 40 epochs in total for all pretraining losses. 
    C. TR@1: Text Retrieval at top-$1$ on COCO; F. TR@1: Text Retrieval at top-$1$ on Flickr. C. Caption: captioning performance in CIDEr on the COCO dataset.
    VQA is on \texttt{test-dev}. NOCAPS is on validation in terms of CIDEr.
    Momentum teacher network is disabled. The highest number for each task is bolded.    }
    \label{tbl:varying_pre_40epoch_total}
\end{table*}

\noindent\textbf{Weight of Distillation Losses.}
In the pretraining, we add the distillation loss.
Table~\ref{tab:dist_weight_40epochs} shows the experiments with different weights on the loss
items with 40 epochs. With a positive weight, the zero-shot retrieval task is improved about 1 to 3 points, but after the fine-tuning, the improvement vanishes.
On VQA, it gives non-trivial $0.1\sim 0.4$ points' improvement. On COCO captioning task, we can observe about 1 points improvement when the weight is $0.6$ or $1.0$.
Overall, we consider the optimal weight is around $0.6\sim 1.0$.
Table~\ref{tab:dist_weight_80epochs} shows the results with 80 pretraining epochs. As we can see, 1.0 gives the best accuracy over all tasks, and we also apply 1.0 for other
network structures. 

\begin{table*}[]
    \centering
    \small
    \begin{tabular}{ccccccc}
    \toprule
    \multirow{2}{*}{weight}  & \multicolumn{3}{c}{Retrieval Task}   & VQA                & \multicolumn{2}{c}{Captioning} \\
                     \cmidrule(lr){2-4}                     \cmidrule(lr){5-5}   \cmidrule(lr){6-7}
                             & ZS F. TR@1  & F. TR@1     & C. TR@1  & \texttt{test-dev}  & COCO        & \nocaps           \\
    \midrule          
    0                        & 68.7        & 87.2        &\bf69.1   & 71.39              & 116.7       & 74.77            \\% 10M_BASE_FULL_LOSS_SMOOTH_CLIP_WEIGHT1_TEMPERATURE
    0.4                      & 71.5        & 86.5        & 67.1     & 71.43              & 116.4       & 74.44            \\% 10M_BASE_DISTILL_CILP_LEARNABLE_0_4
    0.6                      & \bf72.2     & \bf87.6     &68.9      & 71.58              & \bf118.1    & 74.65            \\% 10M_BASE_DISTILL_CILP_LEARNABLE_0_6, 87.6 for flickr, 68.9 for coco
    0.8                      & 70.2        & 87.4        & 67.7     & \bf71.77           & 116.7       & \bf75.24            \\ % 10M_BASE_DISTILL_CILP_LEARNABLE_0_8
    1.0                      & 69.5        & 87.3        & 68.5     & 71.75              & 117.5       & 74.11            \\ % 10M_BASE_DISTILL_CILP_LEARNABLE_1_0
    \bottomrule
    \end{tabular}
    \caption{Different weights on the distillation loss in pretraining with 40 epochs. Retrieval task and COCO captioning task are on test split while \nocaps is on validation split.}
    \label{tab:dist_weight_40epochs}
\end{table*}

\begin{table*}[]
    \centering
    \small
    \begin{tabular}{ccccccc}
    \toprule
    \multirow{2}{*}{weight}  & \multicolumn{3}{c}{Retrieval Task}   & VQA                & \multicolumn{2}{c}{Captioning} \\
                     \cmidrule(lr){2-4}                     \cmidrule(lr){5-5}   \cmidrule(lr){6-7}
                             & ZS F. TR@1  & F. TR@1     & C. TR@1  & \texttt{test-dev}  & COCO        & \nocaps           \\
    \midrule     
    0.6              &  69.4          & 89.0    & 71.3       &  72.26      &  118.9       & 77.07   \\ % 10M_BASE_DISTILL_CILP_LEARNABLE_0_6_80E
    0.8              &  \bf71.6          & 88.8    & 71.6       &  72.24      &  \bf119.4       & 76.55   \\ % 10M_BASE_DISTILL_CILP_LEARNABLE_0_8_80E
    1.0              &  \bf71.6          & \bf89.9    & \bf72.1       &  \bf72.42      &  \bf119.4       & \bf77.53   \\ % 10M_BASE_DISTILL_CILP_LEARNABLE_1_0_80E 
    \bottomrule
    \end{tabular}
    \caption{Different weights on the distillation loss in pretraining with 80 epochs. 
    Retrieval task and COCO captioning task are on test split while nocaps is on validation split.}
    \label{tab:dist_weight_80epochs}
\end{table*}

\subsubsection{Temperature in Image-Text Contrastive loss}
\begin{figure}
    \centering
    \includegraphics[width=0.95\linewidth]{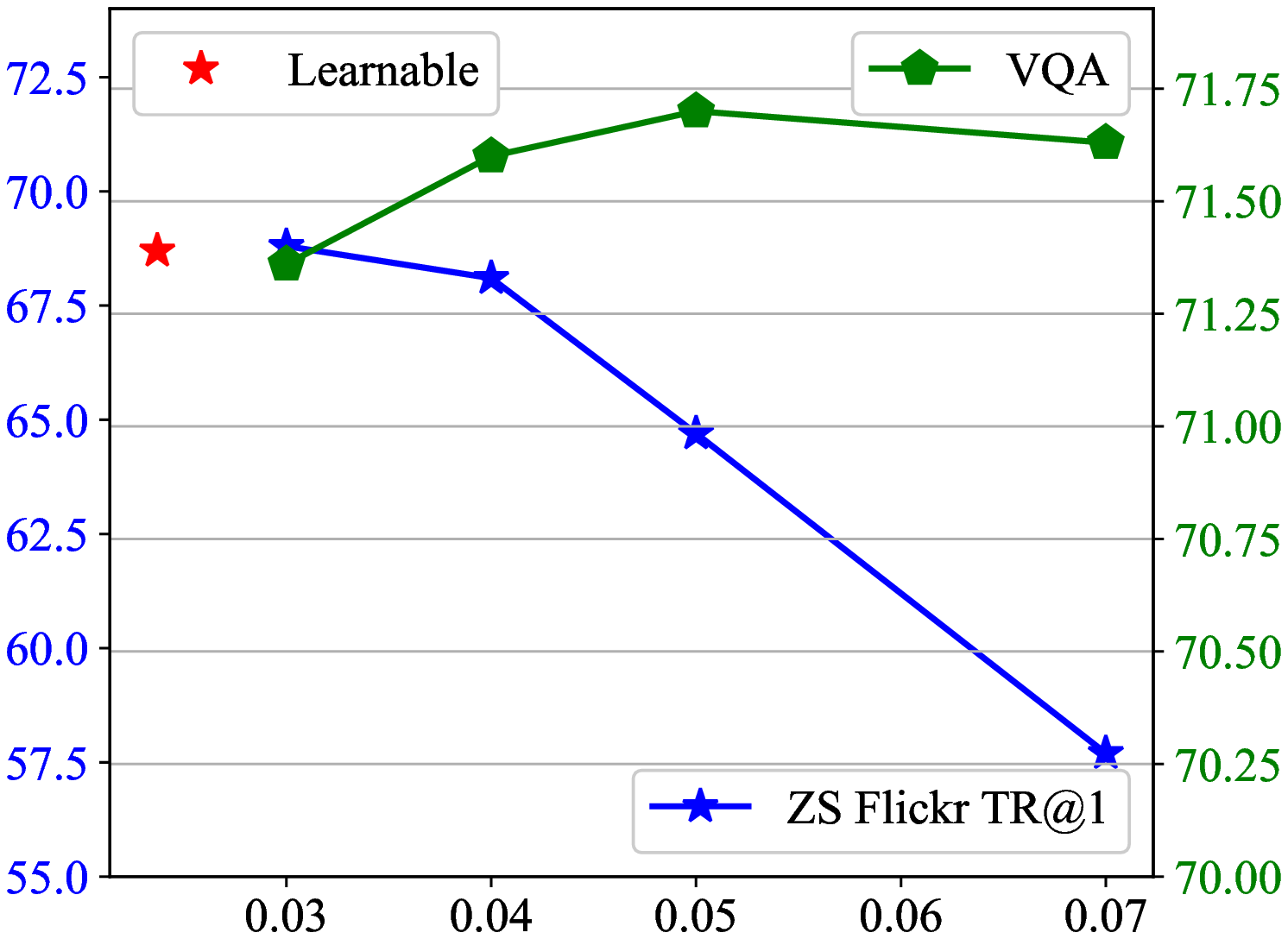}
    \caption{Different ways to set the temperatures in the image-text contrastive loss, including the manually tuned values and the learnable.
    For the learnable, the final learned value is shown in the figure as a red star.
    The left $y$-axis represents the text retrieval at the top-$1$ on Flickr in a zero-shot setting after pretraining. The right $y$-axis represents the fine-tuned VQA performance on
    \texttt{test-dev}.
    % 10M_BASE_FULL_LOSS_SMOOTH_CLIP_WEIGHT1_TEMPERATURE
    }
    \label{fig:temperature}
\end{figure}

In the 
image-text contrastive loss, 
we set the 
temperature learnable as in~\cite{JiaYXCPPLSLD21}.  
Fig.~\ref{fig:temperature} shows the comparison among different ways to set the temperature, including the manually tuned values and the learnable results. 
With an improper pre-defined temperature, the accuracy on the retrieval drops a lot. The VQA performance is relatively more stable. 
When the temperature is learnable, it can achieve a strong retrieval performance of $68.7\%$ and a descent performance on the VQA, and thus we always set the temperature as a learnable parameter.
% ------------------------- keep the following as this is the one for the figure of fig:temperature
% \begin{table*}[]
%     \centering
%     \begin{tabular}{ccc}
%     \toprule
%     \midrule     
%     Learn  & 68.7        & 71.39 \\
%     0.03   & 68.8        & 71.36 \\
%     0.04   & 68.1        & 71.60 \\
%     0.05   & 64.7        & 71.70 \\
%     0.07   & 57.7        & 71.63 \\
%     \bottomrule
%     10M_BASE_FULL_LOSS_SMOOTH_CLIP_WEIGHT1_TEMPERATURE
%     \end{tabular}
%     \caption{Different temperatures in the image-text contrastive loss.
%     Retrieval task and COCO captioning task are on test split while nocaps is on validation split.}
%     \label{tab:temperature}
% \end{table*}

\begin{table}[]
    \centering
    \small
    \begin{tabular}{ccc}
    \toprule
    Loss                     &  VQA   & ZS Flickr TR@1 \\ % & 0-coco-tr \\
    \midrule%
    $l_{\text{m-ITC}}$       & 70.49  & 61.6 \\      % & 50.4      \\% 10M_BASE_EMA_CLIP_LEARNABLE_T
    $l_{\text{ITC}}$         & 71.39  & 68.7 \\      % & 58.7      \\% 10M_BASE_FULL_LOSS_SMOOTH_CLIP_WEIGHT1_TEMPERATURE
    \bottomrule
    \end{tabular}
    \caption{Comparison between the in-batch image-text contrastive loss $l_{\text{ITC}}$
    and the momemtum-based image-text contrastive loss $l_{\text{m-ITC}}$.
    }
    \label{tab:m_itc_vs_itc}
\end{table}

\subsubsection{Momentum-based Image-Text Contrastive Loss}
For the image-text contrastive loss, our implementation $l_{\text{ITC}}$ is based on the in-batch samples, where the negative samples are from the current batch, as in~\cite{RadfordKHRGASAM21,JiaYXCPPLSLD21}.
In~\cite{abs-2107-07651}, the image-text contrastive loss is implemented 
with a momentum encoder~\cite{abs-1911-05722}, which is a drop-in replacement of $l_{\text{ITC}}$. The benefit is that the number of negative samples is independent with the batch size and can be very large, \eg $65536$ in~\cite{abs-2107-07651,abs-1911-05722}. However, the negative samplers are calculated with the momentum encoder which is less accurate. 
We use $l_{\text{m-ITC}}$ to denote this momentum-based loss. 

The temperature is also learnable in $l_{\text{m-ITC}}$ for a fair comparision and the result is shown in Table~\ref{tab:m_itc_vs_itc}.
As we can see, $l_{\text{ITC}}$ achieves better performance on both VQA and the retrieval task in our setting.

\subsubsection{Representative Text Token in Image-Text Contrastive Loss}
The image input contains only one special \texttt{[CLS]} token, while the text contains two special tokens: \texttt{[CLS]}
and \texttt{EOS}. As the text \texttt{[CLS]} is used in the ITM loss, we use the 
\texttt{[EOS]} token in the ITC loss. 
Table~\ref{tab:text_token_in_itc}
shows the comparison of which special text token is used in the ITC loss.
Compared with \texttt{CLS}, the \texttt{EOS} token achieves slightly better accuracy in VQA, but worse performance
in the retrieval task. Thus, we conclude that both performs similarly and in all other experiments, we always use \texttt{[EOS]}
to represent the text input for ITC.

\begin{table}[]
    \centering
    \small
    \begin{tabular}{cccc}
    \toprule
     Text token   &  VQA       & ZS Flickr TR@1 \\ % & 0-Coco-TR \\
    \midrule
    % 10M_BASE_FULL_LOSS_SMOOTH_CLIP_WEIGHT1_CLIP_FIRST
    \texttt{[CLS]}                 & 71.26      & 70.9    \\ %    & 57.9 \\
    % 10M_BASE_FULL_LOSS_SMOOTH_CLIP_WEIGHT1_TEMPERATURE
    \texttt{[EOS]}                  & 71.39      & 68.7    \\ %    & 58.7 \\
    \bottomrule
    \end{tabular}
    \caption{Comparision between which text token should be used in the image-text contrastive loss.}
    \label{tab:text_token_in_itc}
\end{table}

% \subsection{Momentum teacher in fine-tuning}
% 
% \begin{table*}[]
%     \centering
%     \begin{tabular}{ccccccccccc}
%     \toprule
%     \multirow{2}{*}{Input} & w/ Momemtum  &  VQA           & Caption & NOCAPS & Flickr        & Coco-TR       & \multicolumn{2}{c}{NLVR2}         & \multicolumn{2}{c}{SNLI-VE} \\
%                            &              &  test-dev      & test    &  val   & TR@1           & TR@1         & val            & test-P           & val     & test     \\
%     \midrule
%     % 10M_BASE_DISTILL_CILP_LEARNABLE_1_0_80E
%     \multirow{2}{*}{384}  &              & 72.42          & 119.4   & 77.53  & 88.8           & 72.1          & 76.2          & 76.0             &  77.5   &  77.0    \\
%                           &              & 72.42          & 119.9   & 77.36  & 89.2           & 72.0          & 76.2          & 76.1             &  77.8   &  77.1    \\
%     \midrule 
%     \multirow{2}{*}{Large}&              & 74.21          &         &        &                &               &               &                  &         &          \\
%                           &              &                &         &        &                &               &               &                  &         &          \\
%     \bottomrule
%     \end{tabular}
%     \caption{Impact of different input sizes in finetuning. The input size in pretraining is $384$ for all settings.}
%     \label{tab:my_label}
% \end{table*}

\subsubsection{Modal-Type Embedding}
Table~\ref{tab:modal_type_embedding_removal} shows the result by removing the modal-specific embedding.
From the result, the removal of the modal-specific embedding shows no improvement and thus we always add it in all other settings.

% --------------------------------------- can be put in supplementary
% \begin{table}[]
%     \centering
%     \begin{tabular}{cccc}
%     \toprule
%      Modal-type embedding   &  VQA       & 0-Flickr-TR & 0-Coco-TR \\
%     \midrule
%     % 10M_BASE_DISTILL_CILP_LEARNABLE_0_4_DISABLE_TOKEN
%     Yes                & 71.29      & 70.7        & 58.6 \\
%     % 10M_BASE_DISTILL_CILP_LEARNABLE_0_4
%     No                 &  71.43     & 71.5        & 59.7\\
%     \bottomrule
%     \end{tabular}
%     \caption{Comparison of the models with or without the modal-type embedding. The weight of the momentum teacher is $0.4$.}
%     \label{tab:modal_type_embedding_removal}
% \end{table}

\begin{table}[]
    \centering
    \small
    \begin{tabular}{cccc}
    \toprule
     Embedding   &  VQA       & ZS Flickr TR@1 \\ % & 0-Coco-TR \\
    \midrule
    % 10M_BASE_DISTILL_CILP_LEARNABLE_0_6_DISABLE_TOKEN
    No                & 71.33      & 69.7      \\  % & 58.2 \\
    % 10M_BASE_DISTILL_CILP_LEARNABLE_0_6
    Yes               &  71.58     & 72.2      \\  % & 59.6\\
    \bottomrule
    \end{tabular}
    \caption{Comparison of the models with or without the modal-type embedding. }
    \label{tab:modal_type_embedding_removal}
\end{table}

{
\small
\bibliographystyle{ieee_fullname}
\bibliography{egbib}
}

\end{document}